%% file: main.tex
\documentclass[sigconf, screen, nonacm]{acmart}
\AtBeginDocument{%
  \providecommand\BibTeX{{%
    \normalfont B\kern-0.5em{\scshape i\kern-0.25em b}\kern-0.8em\TeX}}}

\setcopyright{cc}
\copyrightyear{2024}
\graphicspath{{./images/}} 

\usepackage{todonotes}
\usepackage{graphicx}
\usepackage{subcaption}
\usepackage{algorithm}

\usepackage{amsfonts}
\usepackage{appendix}
\usepackage[noend]{algpseudocode}

\begin{document}

\title[Sparse vs Contiguous]{Sparse vs Contiguous Adversarial Pixel Perturbations in Multimodal Models: An Empirical Analysis}

\author{Cristian-Alexandru Botocan}

\email{cristian-alexandru.botocan@epfl.ch}
\affiliation{%
  \institution{EPFL}
  \city{Lausanne}
  \country{Switzerland}
}

\affiliation{%
  \institution{armasuisse S+T}
  \city{Thun}
  \country{Switzerland}
}

\author{Raphael Meier}
\authornote{Shared senior authorship}
\email{raphael.meier@armasuisse.ch}
\affiliation{%
  \institution{armasuisse S+T}
  \city{Thun}
  \country{Switzerland}
}
\author{Ljiljana Dolamic}
\authornotemark[1]
\email{ljiljana.dolamic@armasuisse.ch}
\affiliation{%
  \institution{armasuisse S+T}
  \city{Thun}
  \country{Switzerland}
}


\begin{abstract}
Assessing the robustness of multimodal models against adversarial examples is an important aspect for the safety of its users. We craft $L_{0}$-norm perturbation attacks on the preprocessed input images. We launch them in a black-box setup against four multimodal models and two unimodal DNNs, considering both targeted and untargeted misclassification. Our attacks target less than 0.04\% of perturbed image area and integrate different spatial positioning of perturbed pixels: sparse positioning and pixels arranged in different contiguous shapes (row, column, diagonal, and patch). To the best of our knowledge, we are the first to assess the robustness of three state-of-the-art multimodal models (ALIGN, AltCLIP, GroupViT) against different sparse and contiguous pixel distribution perturbations. The obtained results indicate that unimodal DNNs are more robust than multimodal models. Furthermore, models using CNN-based Image Encoder are more vulnerable than models with ViT---for untargeted attacks, we obtain a 99\% success rate by perturbing less than 0.02\% of the image area.
\end{abstract}



\maketitle

\section{Introduction}
\label{sec:intro}
\input{content/intro}

\section{Related Works}
\label{sec:related_works}
\input{content/related_works}

\input{content/method}

\section{Experiments}
\label{sec:experiments}
\input{content/experiments}

\section{Discussion}
\label{sec:discussion}
\input{content/discussion}

\section{Conclusion and Future Work}
\label{sec:conlusion}
\input{content/conclusion}


\bibliographystyle{ACM-Reference-Format}

\newpage
\appendix
\input{content/appendix}
\end{document}

%% file: content/intro.tex
In computer vision, the researchers try to find new methods for improving image classification and other tasks beyond that (e.g. given a set of textual descriptions and an image, the model chooses the most likely image description). For that, the research community developed multimodal models~\cite{radford2021learning, haas2023learning, jia2021scaling, chen2022altclip, li2022blip, xu2022groupvit, pham2023combined, zhong2022regionclip}, which are used as a basis for further research (CLIP~\cite{radford2021learning} led e.g. to various modifications such as RegionCLIP~\cite{zhong2022regionclip}) and are increasingly also considered for commercial usage~\footnote{Deloitte report: \href{https://www2.deloitte.com/uk/en/pages/deloitte-analytics/articles/novel-design-classification-with-clip.html}{Novel Design Classification with CLIP}}. Thus, it is essential to use such models, knowing their robustness and security implications. While prior studies have demonstrated the vulnerability of multimodal models to perturbations of the whole image~\cite{mao2022understanding,qiu2022multimodal}, their vulnerability to pixel-level perturbations remains largely unexplored. In particular, the impact of the number of perturbed pixels and their spatial distribution on the attack performance has not been explored so far. Both of these basic aspects must be considered when constructing adversarial attacks using pixel perturbations. Empirical analysis of their impact on attack performance would enable more informed design choices when crafting novel adversarial pixel perturbations and ultimately lead to a better understanding of the overall robustness of multimodal models.

In this work, we develop methods and experiments to address this gap. We focus on the black-box scenario because we want to confront the security problem of an attacker who does not have any prior information about the model. We rely on $L_{0}$-norm perturbations to have control over the number of pixels being perturbed. We extend \emph{Sparse} pixel distribution attack presented in work by Su et al.~\cite{su2019one} to incorporate different spatial encodings resulting in five different \emph{Contiguous Attacks} (details are described in Section~\ref{sec:contiguous}). Depending on the number and distribution of perturbed pixels, our method exploits potential vulnerabilities hidden in the model architectures, such as the kernel sizes in the case of CNN-based Image Encoder of multimodal models. Pixel perturbations are usually performed in the original image. This assumes a threat model with minimal information available to the adversary (i.e., does not need to know the preprocessing routine). However, if you perturb the preprocessed image, you can be certain that your experimental results are not confounded by the different preprocessing pipelines, which we prioritized for this study. We evaluate the effectiveness of these attacks with images from ImageNet on four open-source models (ALIGN~\cite{jia2021scaling}, AltCLIP\cite{chen2022altclip}, CLIP-B/32~\cite{radford2021learning}, and GroupViT~\cite{xu2022groupvit}) representative of the domain. Additionally, we evaluate our approach for two state-of-the-art Deep Neural Networks (DNNs), which are trained on the ImageNet dataset~\cite{krizhevsky2012imagenet}. Finally, we also release the code for reproducing the study: \href{https://github.com/ChristianB024/SparseVsContiguityRepo}{https://github.com/ChristianB024/SparseVsContiguityRepo}

Our empirical analysis led to several key insights:
\begin{itemize}
\item We found that three out of four multimodal models and both unimodal DNN models are most vulnerable to the \emph{Sparse Attack} for all examined attack scenarios (targeted and untargeted misclassification).
\item The \emph{Patch Attack} is most effective against the CNN-based multimodal model (ALIGN ~\cite{jia2021scaling}). In an untargeted scenario, we reach a 99\% success rate only by perturbing 0.01915\% of the image (16 pixels).
\item Overall, we observe that the multimodal models are more vulnerable to pixel perturbation attacks than the state-of-the-art DNNs, which we suspect to be linked to the way multimodal models and unimodal DNNs are trained.
\end{itemize}

%% file: content/related_works.tex
\subsection{Advesarial Attacks on multimodal models}
In recent years, multimodal models~\cite{radford2021learning, haas2023learning, jia2021scaling, chen2022altclip, li2022blip, xu2022groupvit, pham2023combined, zhong2022regionclip}, which combine information from different modalities such as text, image, and audio, have emerged rapidly as powerful tools across various applications ranging from natural language processing and computer vision to speech recognition. Previous research studied attacking techniques on multimodal models and tried to identify certain vulnerabilities in those models. 

An adversarial perturbation attack can be defined by the approach used for perturbing the input. For instance, the works~\cite{cao2023less, fort2021pixels} explored the CLIP~\cite{radford2021learning} multimodal model's robustness against typographical attacks---the attacker introduced a textual sticker containing a word on the image to misclassify the image to the class written on the textual sticker.

The paper by Qiu et al.~\cite{qiu2022multimodal} analyzed the robustness of multimodal models by adding noise/blur to the whole image to create the image perturbations, while in our work, we focus on perturbing only a limited number of pixels. Similarly, the study by Mao et al.~\cite{mao2022understanding} quantified the robustness of the CLIP model by creating adversarial images using white-box PGD attack~\cite{madry2017towards}. They also presented two defense methods for mitigating that kind of attack. Moreover, some works, such as Yang et al.~\cite{yang2021defending}, focus on adversarial perturbation against multimodal models using only the PGD white-box scenario. Similarly, in the work by Schlarmann et al.~\cite{schlarmann2023adversarial}, they launch imperceptible attacks using the same PGD white-box technique on a single open-source model (OpenFlamingo~\cite{awadalla2023openflamingo}). In contrast,  in our study, we limit the adversary's power to considering the black-box scenarios only. A more different attack is presented in the work by Freiberger et al.~\cite{freiberger2023clipmasterprints} where it exploited the vulnerability of the modality gap in a contrastive learning approach by generating adversarial examples using the evolution strategy for searching generative models' latent~space.
\subsection{Adversarial attacks using Genetic Algorithms (GA)}
Certain adversarial perturbation attacks on Deep Learning models focused only on the white-box scenarios~\cite{szegedy2013intriguing,goodfellow2014explaining, kurakin2016adversarial, madry2017towards, papernot2016limitations, carlini2017towards, moosavi2016deepfool, athalye2018obfuscated}, while others evolved in direction of black-box attacks~\cite{su2019one, chen2017zoo, brendel2017decision, tu2019autozoom}. Moreover, for improving their attacks, there are studies where genetic algorithms have been introduced~\cite{su2019one, wu2021genetic}. For instance, the study by Su et al.~\cite{su2019one} generated the adversarial examples against the DNNs by perturbing a specific number of pixels, thus using an $L_{0}$ norm. Furthermore, the paper by Jere et al.~\cite{jere2019scratch} used evolutionary strategies (including Differential Evolution) for generating \emph{scratch} perturbations on the image against DNNs only. Similarly, in terms of applying the evolutionary algorithms but using sparse adversarial perturbations through simultaneous minimization of $L_{0}$ and $L_{2}$ norms, the research by Williams et al.~\cite{williams2023black} restricted the space of potential perturbations quite aggressively and focused on studying only sparsely distributed pixel perturbations. So there is no exploration regarding the impact of contiguous shapes of the perturbation on attack performance.

In addition, both studies~\cite{su2019one, williams2023black} performed the pixel perturbations on the original images. In this work, we perturb the preprocessed image such that our attack is neither influenced by the dimension of the original image nor the particular preprocessing routines. Therefore, we investigate the attack based solely on the different models. Also, the work by Qiu et al.~\cite{qiu2021black} included an analysis of creating the image perturbations with respect to the $L_{\infty}$-norm in a black-box setup, using different evolutionary strategies and characterizing which evolutionary algorithm is more suitable for this kind of attack. More recently, the paper by Nam et al.~\cite{nam2023aesop} explored the one-pixel idea but using an exhaustive search procedure, and the perturbations were identified directly on the original image while we are attacking the preprocessed images. In the study by Ghosh et al.~\cite{ghosh2022black}, the differential evolution (DE)~\cite{storn1997differential} technique generated the pixel-perturbations in a black-box setup. The attack was launched against relatively old CNN models (VGG16~\cite{simonyan2014very}, GoogleNet~\cite{szegedy2015going}, InceptionV~\cite{szegedy2016rethinking}, ResNet-50~\cite{he2016deep}), while we attack more recently proposed multimodal models. Since we want to compare their robustness with established unimodal deep neural networks, we also compare them against the ResNet-50~\cite{he2016deep} and the VAN (Visual Attention Network~\cite{guo2023visual}, which used Dilated Convolution Layers~\cite{yu2015multi}).

\subsection{Adversarial Patch Attacks}
The idea of perturbing pixels in the form of a patch is not new, and it was already tested before in the white-box~\cite{brown2017adversarial, karmon2018lavan} and black-box setups~\cite{brown2017adversarial, zhou2021data}. For instance, the work by Alrasheedi et al.~\cite{alrasheedi2023imperceptible} explored the problem of making the image perturbations imperceptible and also found certain boundaries in the width limit. However, the entire attack was launched in a white-box setup, while we focus on black-box attacks since they are much closer to real-world scenarios.
The study by Wei et al.~\cite{wei2022simultaneously} performed the attacks in a black-box manner by applying reinforcement learning as an optimization technique to simultaneously find the best position and perturbation values for the patch. This process is similar to what we try to achieve in our experiments, but we use the DE approach to evolve our perturbations. Different methods were used in the context of patch attacks by even hiding the patch as a QR code~\cite{chindaudom2020adversarialqr, chindaudom2022surreptitious}, such that the pixel perturbations became less suspicious to the human eye. Moreover, these patch adversarial attacks were also examined concerning object detection models using aerial images~\cite{tang2023adversarial} and deep hashing models~\cite{hu2021advhash}.

Lastly, the research community relied on GANs to generate adversarial patches for image classification~\cite{ bai2021inconspicuous, demir2018patch, liu2019perceptual} or object-detection models~\cite{hu2021naturalistic, pavlitskaya2022feasibility}. With this technique, the recent paper by Zhou et al.~\cite{zhou2023advclip} showed a non-targeted attack against the multimodal model CLIP~\cite{radford2021learning}. However, we do not train GAN models since we try to limit the attacker's power (including the computational one).

%% file: content/method.tex
\section{Attack model}
\label{sec:threat_model}
\begin{figure*}[tb]
    \centering
    \includegraphics[width=0.70\textwidth]{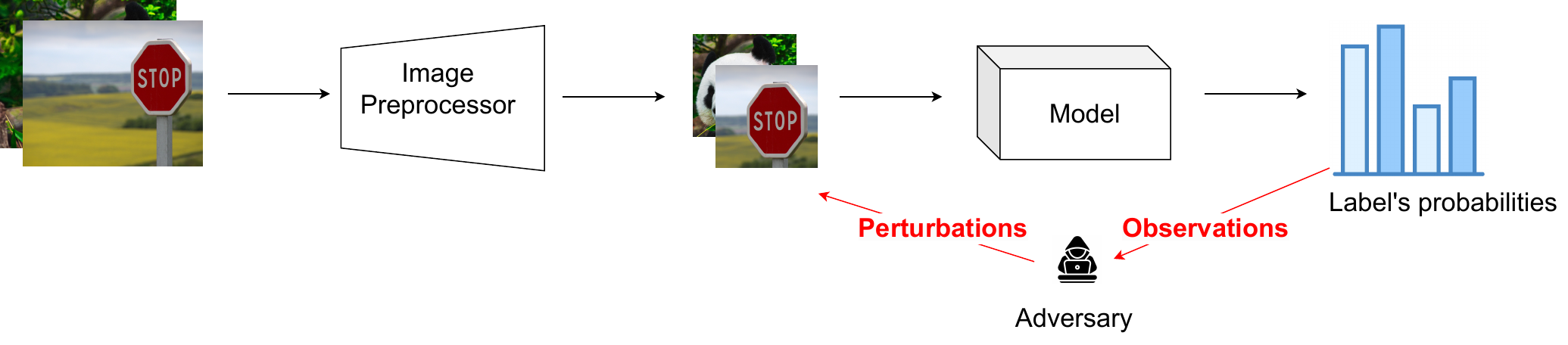}
Model    \caption{Threat Model Visualization}
    \label{fig:threat_model_image}
\end{figure*}
We analyze the black-box scenario of an adversary with access to Artificial Intelligence (AI) models during inference, so the attacker has access only to the model's input and output. Figure~\ref{fig:threat_model_image} presents a visualization of the described threat model. To formalize the attack, let the function \( f : \{0, ..., 255\}^{w \times h \times 3} \rightarrow \mathbb[0, 1]^C \) represent the model classification which takes as an input \( x \in  \{0, ..., 255\}^{w \times h \times 3}\) representing an RGB image with dimension $w$ and $h$ for width and height respectively. The output corresponds to a probability distribution for the $C$ classes. The maximum probability from the list of all given classes $C$ yields the label (or textual description) $y_{original} = \arg\max_{i \in \{1, \ldots, C\}} f_i(\mathbf{x})$. We consider two possible pixel-perturbation attacks. In the first one, the goal of the attacker is to find a perturbation ($\Delta \mathbf{x}$) such that the new image is classified as a specific target label (targeted attack noted as \emph{tar}). The attack is presented in the Equation~\ref{eq:target_for} where the adversary is maximizing the target label ($p_{tar}$) probability by perturbing at most $d$ pixels.

\begin{equation}
\label{eq:target_for}
\begin{aligned}
&\text{find} \; \Delta \mathbf{x} \; \text{such that}
\begin{cases}
    p_{\text{tar}} = \underset{\Delta \mathbf{x}}{\max} \; f_{i:=\text{tar}}(\mathbf{x}+\Delta \mathbf{x}), \\
   \text{and} \; ||\Delta \mathbf{x}||_0 \leq d
\end{cases}
\end{aligned}
\end{equation}
The second one, untargeted attack (noted as \emph{untar}), has as the goal to do random misclassification (maximizing any other class probability) as it is described similarly by the Equation~\ref{eq:untarget_for}. 

\begin{equation}
\label{eq:untarget_for}
\begin{aligned}
&\text{find} \; \Delta \mathbf{x} \; \text{such that}
\begin{cases}
   p_{untar} = \underset {\Delta \mathbf{x}}{\max } \; f_{i\neq original}(\mathbf {x}+\Delta \mathbf{x}), \\ 
   \text{and} \; ||\Delta \mathbf{x}||_0 \leq d
\end{cases}
\end{aligned}
\end{equation}

The perturbation of our attack is created for the preprocessed image instead of the original image. In fact, if the adversary perturbs one pixel in a $32\times32$ image (dimension of the original image), then it perturbed 10.24\% of the image (approach presented in the paper by Su et al.~\cite{su2019one}), while one pixel in a preprocessed image of dimension $224\times224$ represents less than 0.002\% of the image area. Table~\ref{tab:area} shows the exact percentages of the perturbed area for the target models. Furthermore, attacking this small area of the image makes the attack less perceptible for the human eye and hard to detect by the systems, hence having potentially catastrophic consequences in a sensitive application. Moreover, this setup is closer to the real-world scenario, where companies usually release new AI models via an API, which takes input images of a specific dimension (the company also need to release the Image Processor of the model) and outputs the probabilities for specific classes/textual descriptions.

Lastly, the attack is limited to $Q$ number of queries to the model. Hence, the attacker creates the perturbations using an evolutionary algorithm (differential evolution---DE~\cite{storn1997differential}) and makes a maximum of $Q$ queries.

\begin{table*}[tb]
\caption{Percentage of the perturbed area in the attacked image}
\label{tab:area}
\begin{tabular}{|l|l|l|l|}
\hline
Preprocessed image dimension \textbackslash Pixels & 4  pixels       & 9 pixels        & 16 pixels       \\ \hline
ALIGN - 289 x 289                                  & 0.00478\% & 0.01077\% & 0.01915\% \\ \hline
AltCLIP, CLIP-ViT-B/32, GroupViT - 224 x 224 & 0.00797\% & 0.01793\% & 0.03188\% \\ \hline
VAN-base, ResNet-50 - 224 x 224                              & 0.00797\% & 0.01793\% & 0.03188\% \\ \hline
\end{tabular}
\end{table*}

\section{Evolutionary Attacks}
\label{sec:genetical_algo}
This section presents the encoding for each type of attack used (\emph{Sparse} and \emph{Contiguous Attacks}), followed by how we initialize the genetic algorithm setup. The evolutionary process is then explained in depth, covering mutation, crossover, and fitness functions.
\subsection{Sparse Attack}
\label{sec:sparse}
\begin{figure}[tb]
    \centering
    \includegraphics[width=0.27\textwidth]{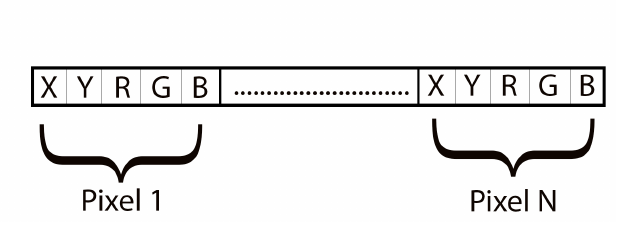}
    \caption{\small{Agent encoding for the \textbf{Sparse Attack}}}
    \label{fig:sparse_encoding}
\end{figure}
As mentioned in Section~\ref{sec:threat_model}, we use Su et al.~\cite{su2019one}'s approach, the differential evolution (DE)~\cite{storn1997differential} genetic algorithm, as an optimization method for finding the best perturbations for misclassification tasks. Encodings of the perturbations in specific vectors (noted as agents or population members) become accessible to the evolutionary approach, which tries to iteratively yield better perturbations over time. For the \textbf{Sparse Attack}, we can use five values, $[x_{coord}, y_{coord}, r_{value}, g_{value}, b_{value}]$, to describe a pixel, namely its positioning by the coordinates and its RGB values. As a result, an agent is represented by $N$ flattened and concatenated vectors, where $N$ is the number of pixels for which the perturbation is being performed. Section~\ref{fig:sparse_encoding} illustrates the encoded perturbation vector we use for this type of attack. An important property of the \emph{Sparse Attack} is that there are no constraints on the spatial distribution of the pixels used for the attack.
\subsection{Contiguous Attacks}
\label{sec:contiguous}
\begin{figure*}[tb]
  \centering
  \begin{subfigure}{0.19\textwidth}
    \centering
    \includegraphics[width=0.6\linewidth]{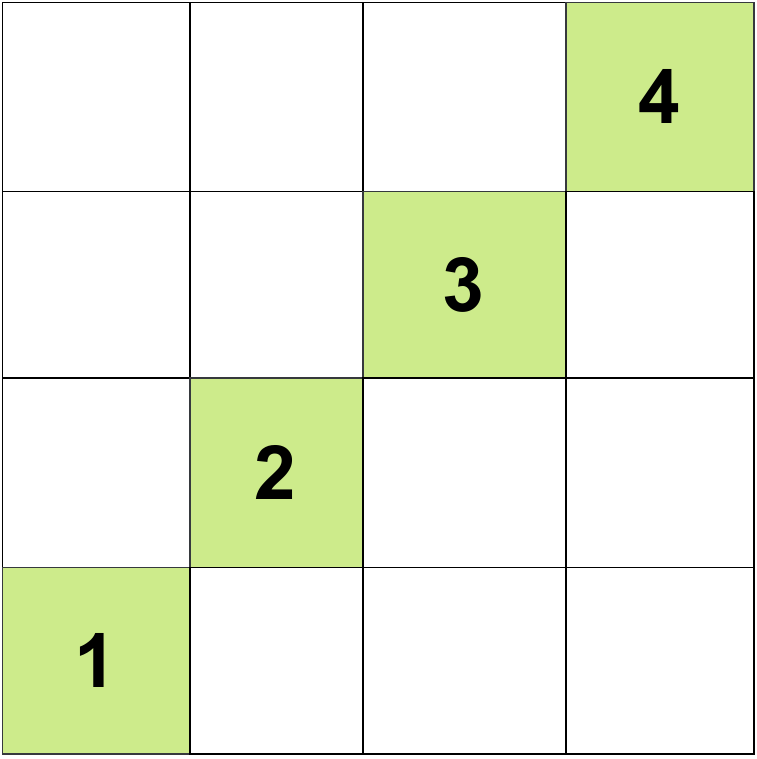}
    \caption{\small{Anti-Diagonal}}
  \end{subfigure}
  \begin{subfigure}{0.19\textwidth}
    \centering
    \includegraphics[width=0.6\linewidth]{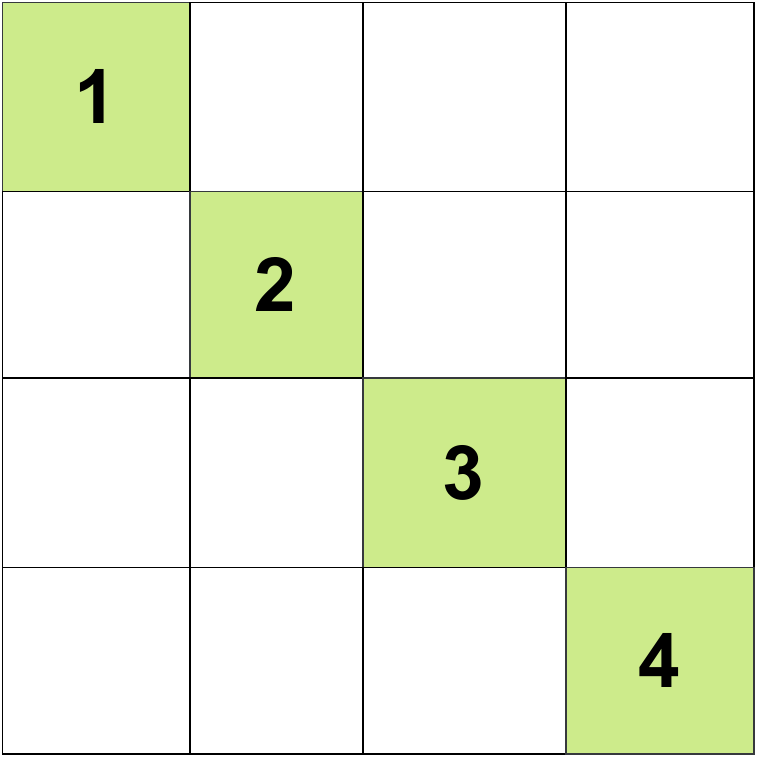}
    \caption{\small{Diagonal}}
  \end{subfigure}
  \begin{subfigure}{0.19\textwidth}
    \centering
    \includegraphics[width=0.6\linewidth]{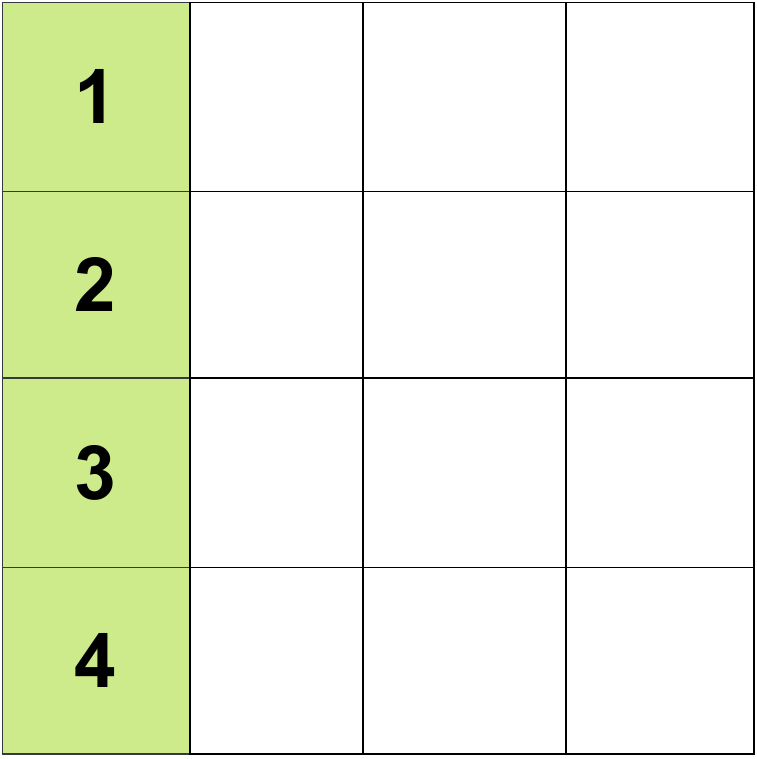}
    \caption{\small{Column}}
  \end{subfigure}
  \begin{subfigure}{0.19\textwidth}
    \centering
    \includegraphics[width=0.6\linewidth]{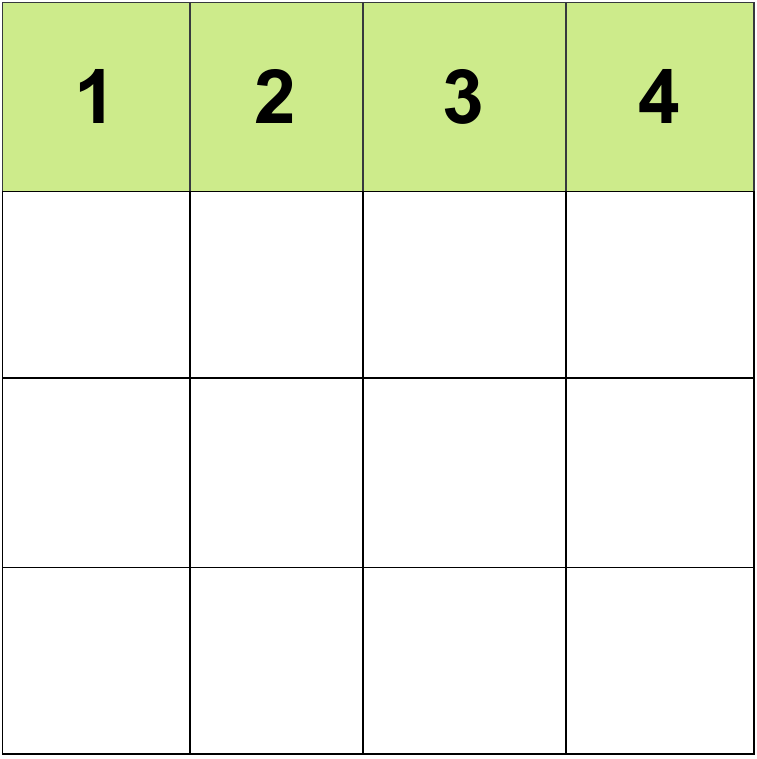}
    \caption{\small{Row}}
  \end{subfigure}
  \begin{subfigure}{0.19\textwidth}
    \centering
    \includegraphics[width=0.6\linewidth]{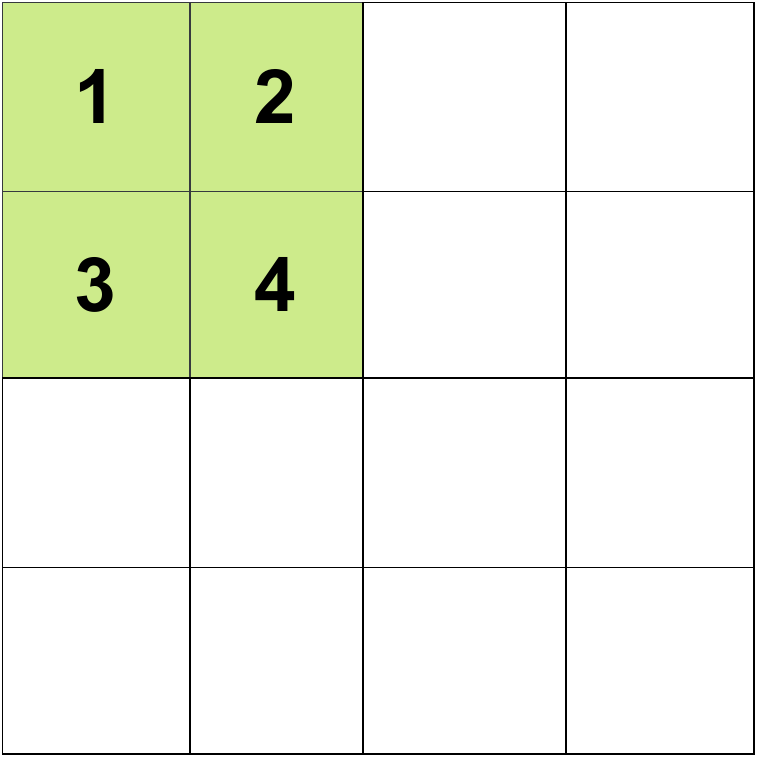}
    \caption{Patch}
  \end{subfigure}
  \caption{\small{Spatial arrangement of pixels in Contiguous Attacks for exemplary case of four perturbed pixels}}
  \label{fig:pixels_cont}
\end{figure*}

In contrast to the \emph{Sparse Attack}, we can impose constraints on the spatial distribution of the pixels used for an attack. In particular, we choose to force pixels to form spatially contiguous regions in the image (i.e., \textbf{anti-diagonal}, \textbf{diagonal}, \textbf{column}, \textbf{row}, \textbf{patch}). Since we also address the question of how different shapes of contiguous pixels can affect the model robustness, we create appropriate encodings for the \emph{Contiguous Attack}. We decrease the number of the values that must be mutated for other shapes and store only the coordinates of the first pixel from that shape. Thus, we have the generic encoding pattern $[x_{coord}, y_{coord}, r_{value_1}, g_{value_1}, b_{value_1},..., r_{value_n}, g_{value_n}, b_{value_n}]$, where the $r$, $g$, $b$ values represent the colors for each perturbed pixel in the specific shape. The first two values are the x and y coordinates, representing different starting pixels:
\begin{description}
      \item[Anti-Diag Attack] the x and y coordinates of the lowest pixel of the shape.
      \item[Diagonal Attack] the x and y coordinates of the up-most pixel of the diagonal.
      \item[Column Attack] the x and y coordinates of the up-most pixel of the column.
      \item[Row Attack] the x and y coordinates of the left-most pixel of the row.
      \item[Patch Attack] the x and y coordinates of the up-left corner pixel of the patch.
\end{description}
As an example on how the pixels are positioned and what is their order in the encoding vector depending on the attack type, we illustrate the case where we perturb four pixels in Figure~\ref{fig:pixels_cont}.

\begin{figure}[tb]
    \centering
    \begin{subfigure}{0.43\textwidth}
        \includegraphics[width=0.95\textwidth]{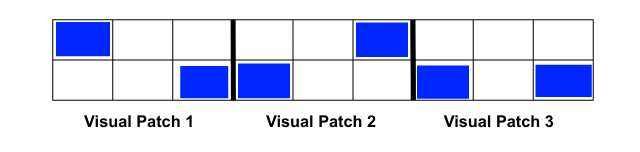}
        \caption{\small{Sparse Attack}}
    \end{subfigure}
    \begin{subfigure}{0.43\textwidth}
        \includegraphics[width=0.95\textwidth]{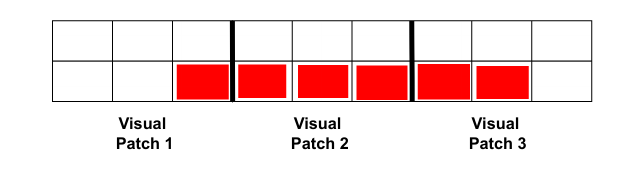}
        \caption{\small{Contiguous - Row Attack}}
    \end{subfigure}
    \hfill
     \begin{subfigure}{0.43\textwidth}
        \includegraphics[width=0.95\textwidth]{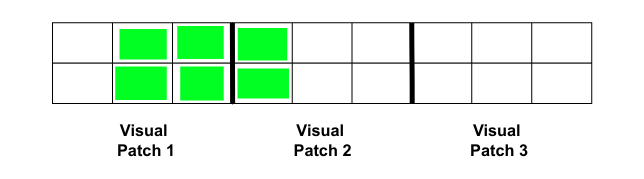}
        \caption{\small{Patch Attack}}
    \end{subfigure}
    \caption{\small{Pixel perturbations for different attacks on patch-based models (e.g. Vision Transformers, ViTs)}}
    \label{fig:all}
\end{figure}

The idea of \emph{Contiguous Attacks} is to actually try to perturb multiple visual patches (depending on the length of the shape) from the ViT model. So, suppose an attacker chooses the row shape. In that case, it can perturb a number of consecutive visual patches (or tokens) differently depending on their alignment with the row of perturbed pixels. Figure~\ref{fig:all} shows an example of perturbing 6 pixels in different shapes. We have most perturbations in the middle visual patch for the \emph{Row Attack} and less for the beginning and final patch. Hence, we want to test if it is better to perturb less information but from many very different visual patches (\emph{Sparse Attack}), perturb less information to some visual patches and more in some others (\emph{Row Attack}), or perturb more in a smaller number of visual patches (\emph{Patch Attack}). Thus, we do not focus on the evolutionary method for improving the position of those pixel shapes. However, we emphasize that if we use an evolutionary approach to find the best perturbation based on the provided shape, we can still exploit how ViT creates the image tokens. 
\subsection{Initialization}
\label{sec:init}
Similarly with the study by Su et al.~\cite{su2019one}, for the initialization process of the population, we do the following:
\begin{itemize}
    \item for $x_{coord}$ and $y_{coord}$ we assign uniformly random integer values based on the dimension of the attacking image (i.e., $U(0, max_{dim})$)
    \item for RGB values, since their values are between 0 and 255, we assign them as a random value from a normal distribution $N(\mu = 128, \sigma = 127)$
\end{itemize}

\subsection{Evolutionary process}
After initializing the first population (generating $\mathcal{P}$ perturbations, also called agents), we calculate the fitness value for each one. This is computed by applying a fitness function. As mentioned in Section~\ref{sec:threat_model}, we have two possible attack scenarios. Hence, we use two hinge loss functions, inspired by the paper by Chen et al.~\cite{chen2017zoo} and exemplified in the Equations~\ref{eq:F_tar} and \ref{eq:F_untar}---for targeted attacks $F_{tar}$ and $F_{untar}$ for the non-targeted attacks. Furthermore, we maximize the fitness function of every new generation. Since we assume a black-box attack scenario, we need to consider that we can only access the probabilities for each class. For instance, in our case, the $p[x]$ means we obtain the softmax (probability) value from the model for class $x$. Also, as these functions ascend, sorting the agents by their fitness values, at position zero in the population yields the best perturbation.
\begin{align}
    F_{tar} &= p_{tar} - \max_{i\neq \text{tar}} \, p_i
    \label{eq:F_tar} \\
    F_{untar} &= \max_{i \neq \text{original}} \, p_i - p_{\text{original}}
    \label{eq:F_untar}
\end{align}

\begin{equation}
\label{eq:mutation}
\begin{aligned}
    agent_{cand} = population_0 + mutation_{rate}\cdot(population_{r1} - population_{r2}) \\
    \text{where $r1$ and $r2$ are random indices that are within the population size} \\
\end{aligned}
\end{equation}
For the next iteration, we create $\mathcal{P}$ other new candidates ($agent_{cand}$). Thus, for each member in the population, we apply the mutation function from Equation~\ref{eq:mutation} and the crossover operation, using the exponential strategy approach (see Appendix~\ref{sec:exp_algo}). A list with the parameter values used in the DE is presented in the Section~\ref{sec:experimental_setup}.

After obtaining the candidates, we perform $\mathcal{P}$ queries to the model and compute the fitness values based on the probabilities obtained. Therefore, we decide which $\mathcal{P}$ agents out of $2\cdot \mathcal{P}$ (old and new agents) survive for the next generation, based on their fitness values---the $\mathcal{P}$ agents with the highest fitness values are selected. We repeat this process $\mathcal{G}$ times, as we are limited to the maximum number of queries we can do. The idea of the evolutionary approach is generating increasingly better perturbations over time and, as a result, have at least one perturbation that categorizes the attack as successful. Further details about how we quantify the success of an attack are explained in Section~\ref{sec:empiric}.

%% file: content/experiments.tex
This section presents the images used for the attack and the values for the parameters used. Moreover, we describe the evaluation metrics and baselines. Lastly, we present our numerical results.
\subsection{Experiment setup}
\label{sec:experimental_setup}
\input{content/experimental_setup}

\subsection{Evaluation Metrics and Baselines}
\label{sec:empiric}
Attack performance is quantified by computing the \emph{SR} (success rate) as an empirical measurement. Depending on the type of attack, we define the success of the attack as follows: targeted attacks involves how many images managed to be misclassified as the target label, while for untargeted scenarios, we count the number of images that are misclassified as a different label than the original one. As mentioned Section~\ref{sec:experimental_setup}, for our setup, it translates to using 100 images and computes how many of them we manage to attack successfully. Then, since the SR is a value between 0 and 1, we divide it by 100 (the number of images). Thus, an SR value of 1 means we create successful perturbations for all images.

Lastly, to investigate the effectiveness of the evolutionary attacks, we create random versions for each type of attack and use them as baselines. These \emph{Random Attacks} use the same parameters as the evolutionary one ($\mathcal{P} = 300$ and $\mathcal{G}= 100$), but the difference is in generating those random agents without an evolutionary approach. At each iteration, we generate $\mathcal{P}$ agents according to their respective encodings presented in Sections~\ref{sec:sparse} and \ref{sec:contiguous}, and the values are random and set based on the initialization values presented in Section~\ref{sec:init}. However, some corrections are applied if the values are outside the ranges (especially for the RGB values).
\subsection{Results}
\label{sec:results}
\input{content/results}

%% file: content/experimental_setup.tex
We evaluate our attacks on samples from ImageNet~\cite{krizhevsky2012imagenet}. It contains images with a variable resolution length and, thus, a different number of pixels. As explained in Section~\ref{sec:threat_model}, we attack the preprocessed image before it is fetched into the model; we do not apply any preprocessing operations to the original images from the dataset to scale all of them to a fixed resolution, $256\times256$ resolution, as it is done in the original paper~\cite{krizhevsky2012imagenet}. The number of labels for this dataset is 1000.
\begin{table}[tb]
  \centering
  \caption{Performance comparison between Multimodal Models and DNNs}
  \label{tab:performance_comparison}
  \begin{tabular}{lccc}
    \toprule
    \textbf{Model} & \textbf{Accuracy} & \textbf{Model Type} \\
    \midrule
    ALIGN~\cite{jia2021scaling} & 0.5442 & Multimodal \\
    CLIP\_ViT-B32~\cite{radford2021learning} & 0.5601 & Multimodal \\
    AltCLIP~\cite{chen2022altclip} & 0.6993 & Multimodal \\
    GroupViT~\cite{xu2022groupvit} & 0.3187 & Multimodal \\
    \midrule
    ResNet-50~\cite{he2016deep} & 0.8087 & DNN \\
    VAN-base~\cite{guo2023visual} & 0.8022 & DNN \\
    \bottomrule
  \end{tabular}
\end{table}

The models used can be split into two branches based on their architecture. In the first category, we have the state-of-the-art multimodal models; in the other, we have the state-of-the-art DNNs, trained on the ImageNet dataset. Table~\ref{tab:performance_comparison} shows the list of all the models and their measured accuracy for 10,000 images from the testing dataset of ImageNet. The motivation behind this assessment is given by the connection between robustness, the architecture used, and the versatility of the models (multimodal models can be applied to different datasets without the need for extensive retraining due to the zero-shot property, while DNNs usually require training on a specific dataset and are optimized for that one only).

In order to use the multimodal models for image classification, we rely on appropriate input captions. In particular, we append the string \emph{"a photo of"} in front of the original label of the image to enable zero-shot image classification.

As observed in Table~\ref{tab:performance_comparison}, the models perform very differently, and we want an extensive analysis of their robustness. Thus, to avoid any possible bias, we extract 100 random correctly classified images for each model. Those are part of the testing dataset; consequently, the models did not see those images during their training phase. Then, after extracting those images, we store the preprocessed version based on each model. Hence, we test different attack setups on the same images to quantify the attack's effectiveness.

Based on empirical observations, we observe that the best parameters for the Differential Evolution are $mutation_{rate} = 0.55$ and $crossover_{rate} = 0.8$. Moreover, the population of $\mathcal{P} = 300$ gives the best results when we limit the number of iterations to $\mathcal{G}= 100$. As a remark from Section~\ref{sec:threat_model}, the total number of queries an attacker needs to do to perturb an image is at most $Q$, which is translated to $\mathcal{P} \cdot \mathcal{G} \leq Q$, which means $300 \cdot 100 \leq 30,000 \leq Q$.

%% file: content/results.tex
\begin{figure*}[tb]
    \centering
    \includegraphics[width=0.73\textwidth]{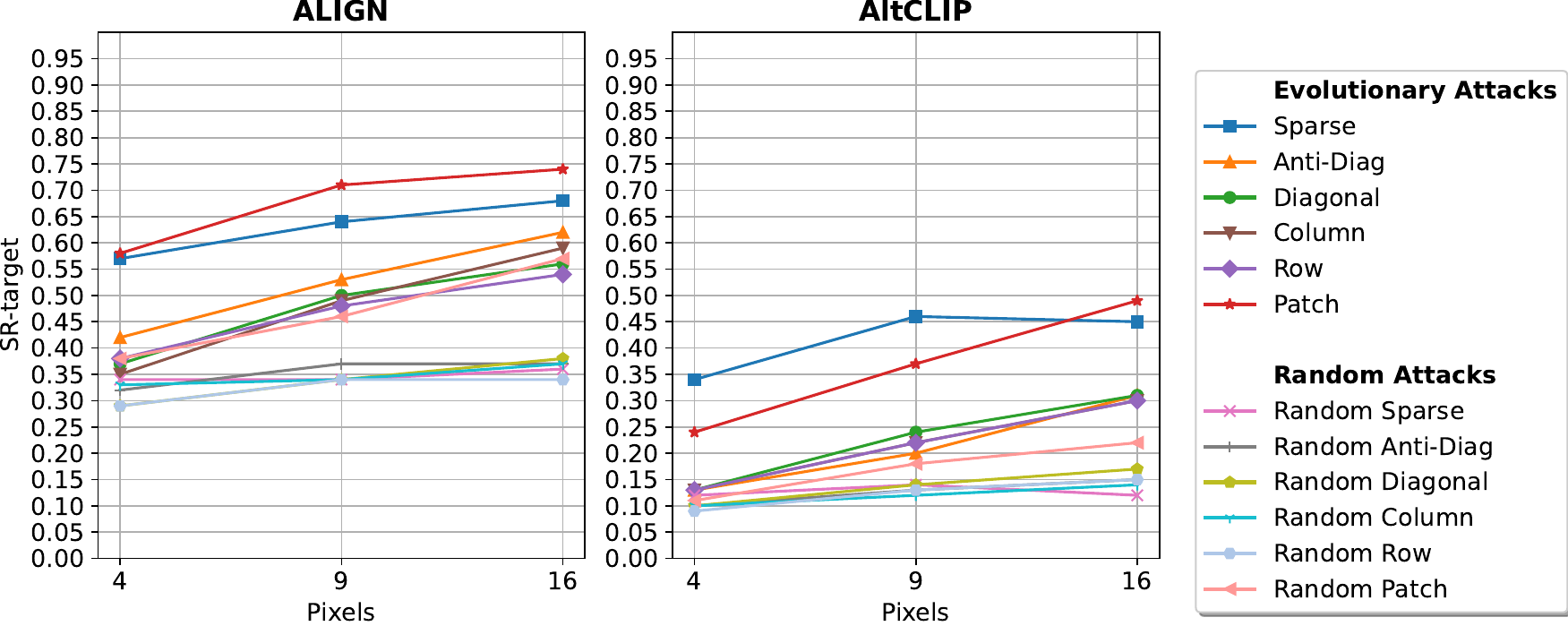}\hfill
    \includegraphics[width=0.73\textwidth]{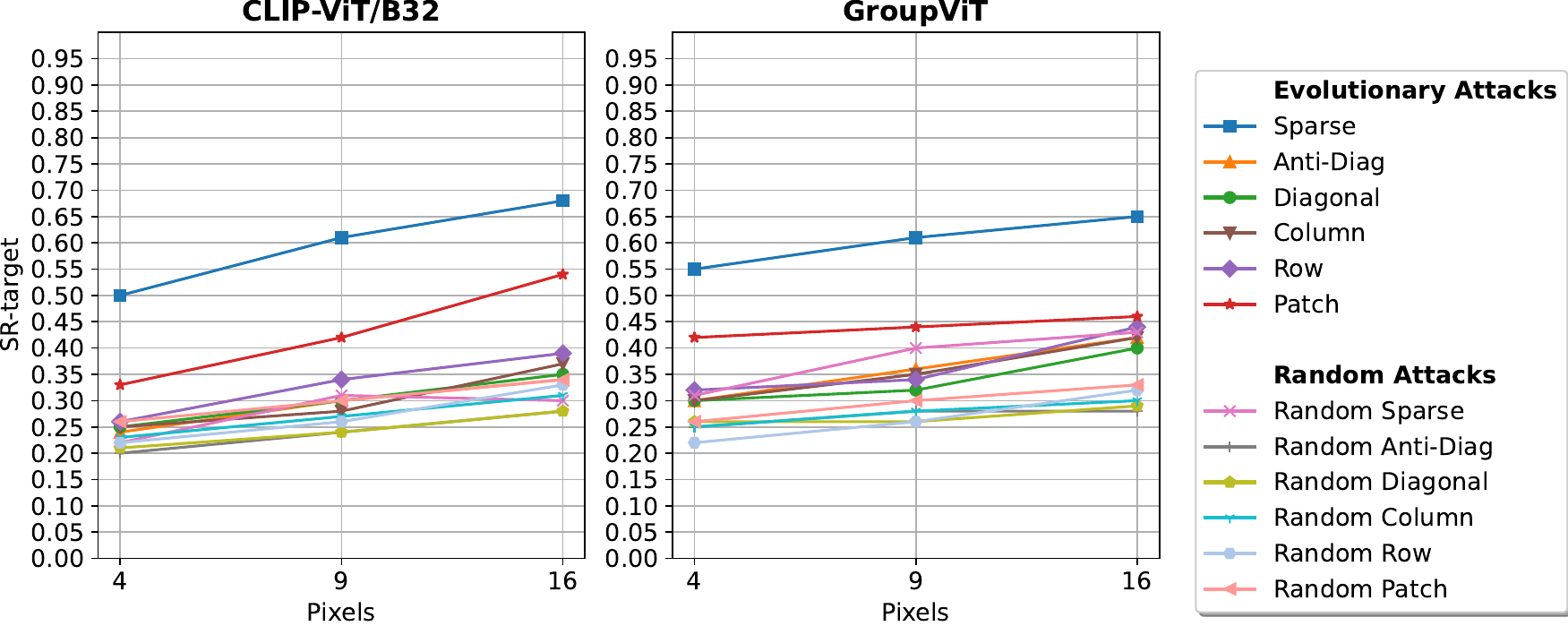}\hfill
    \includegraphics[width=0.73\textwidth]{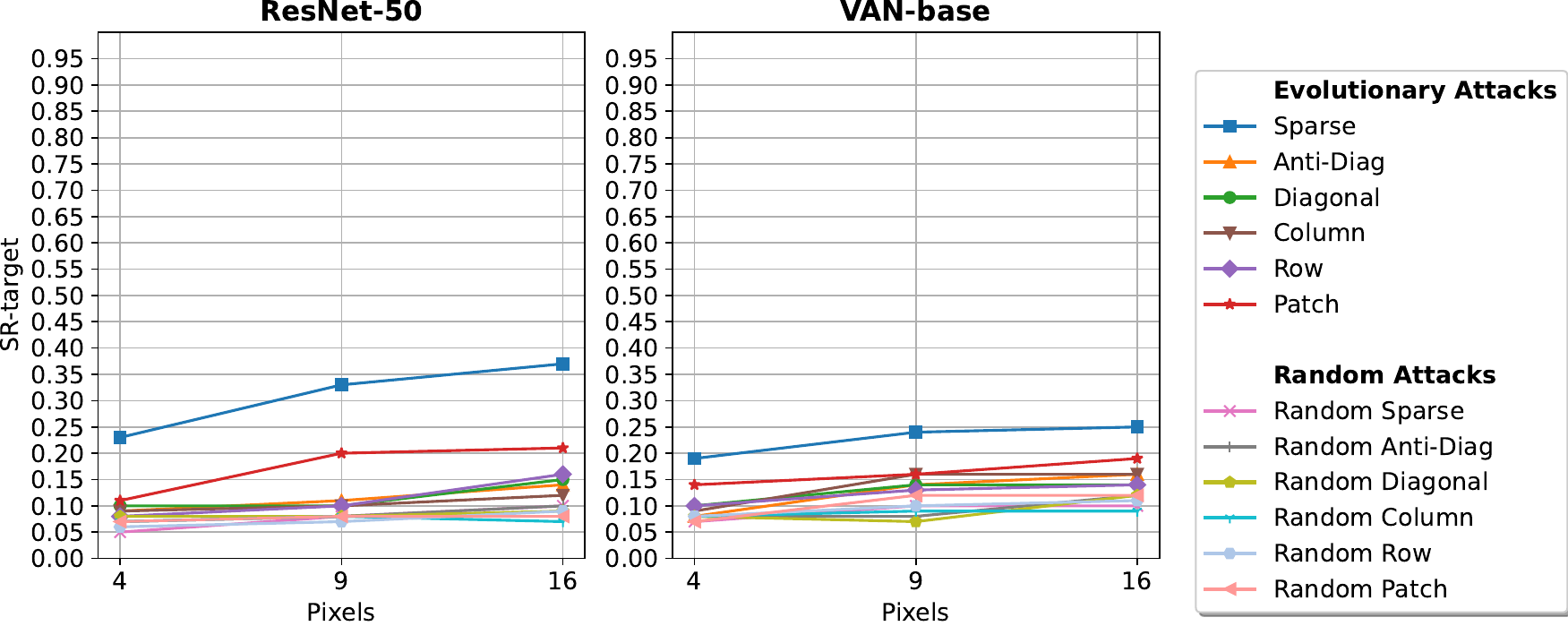}
    \caption{Targeted Attacks}
    \label{fig:tar}
\end{figure*}
\begin{figure*}[tb]
    \centering
    \includegraphics[width=0.73\textwidth]{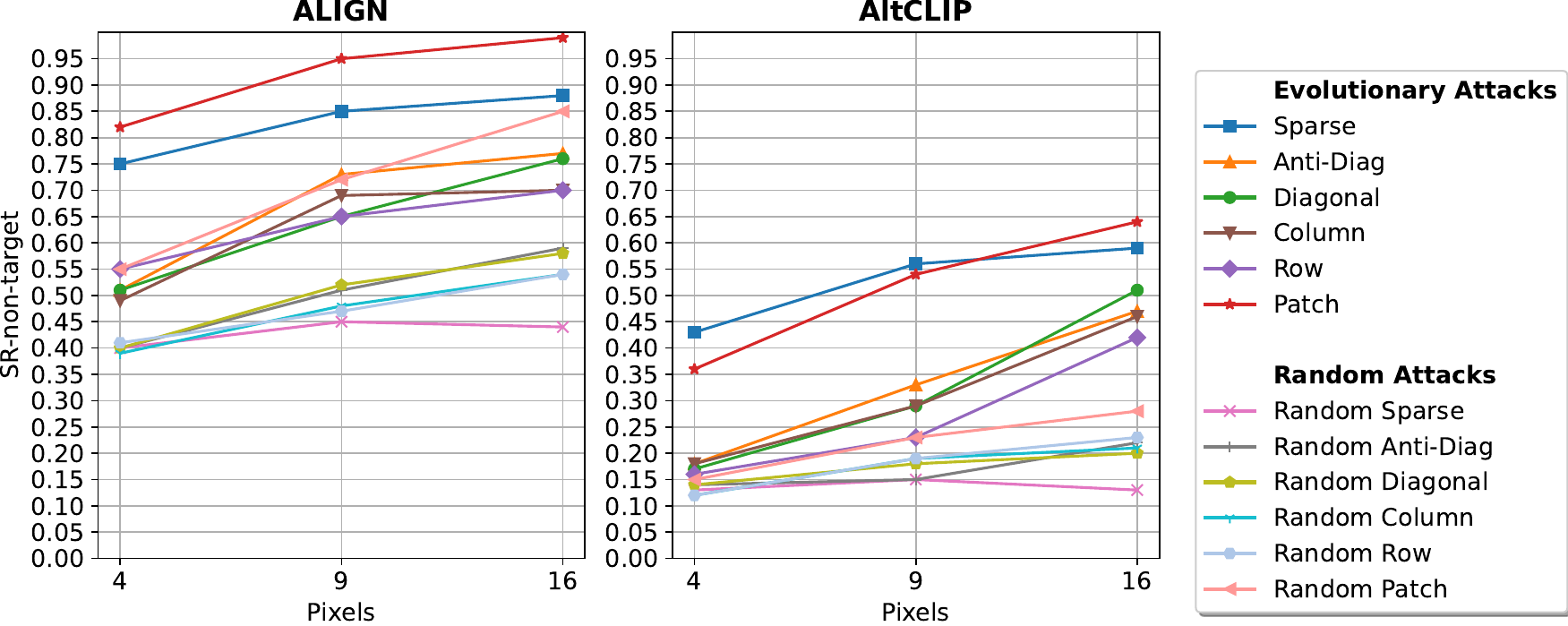}\hfill
    \includegraphics[width=0.73\textwidth]{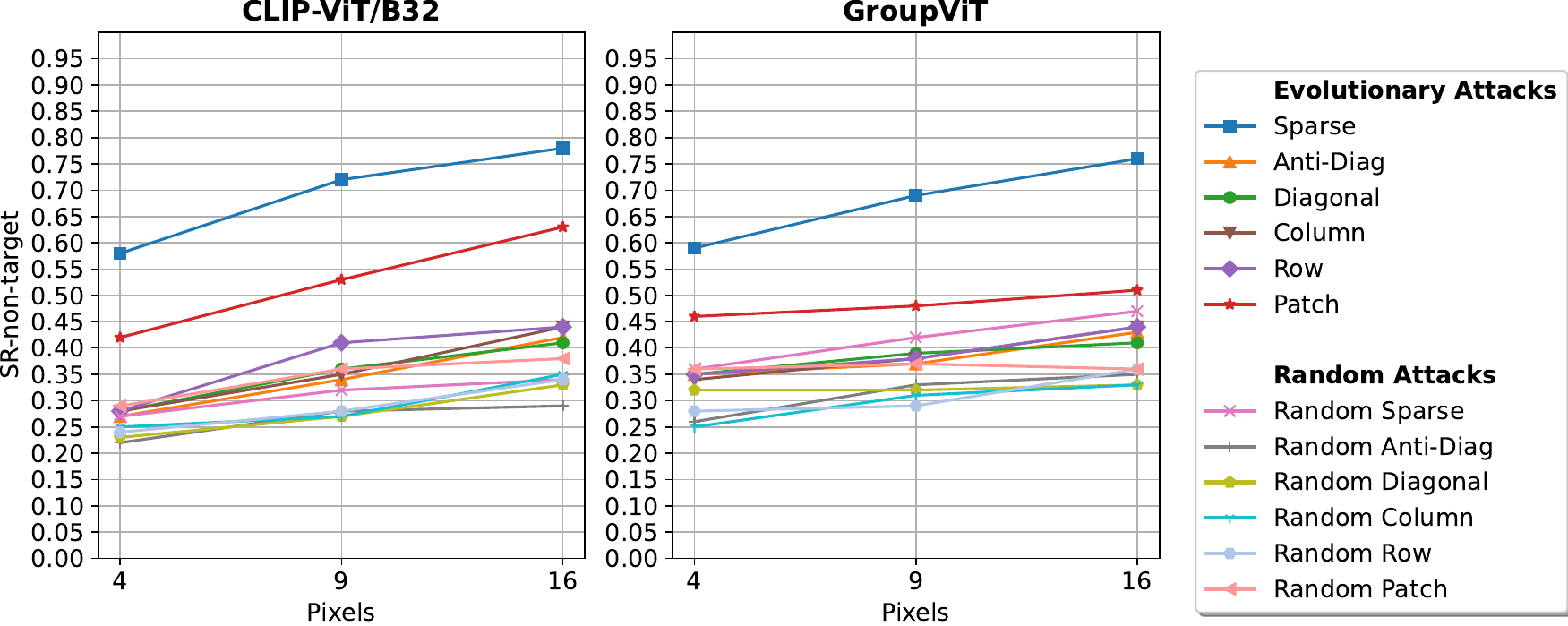}\hfill
    \includegraphics[width=0.73\textwidth]{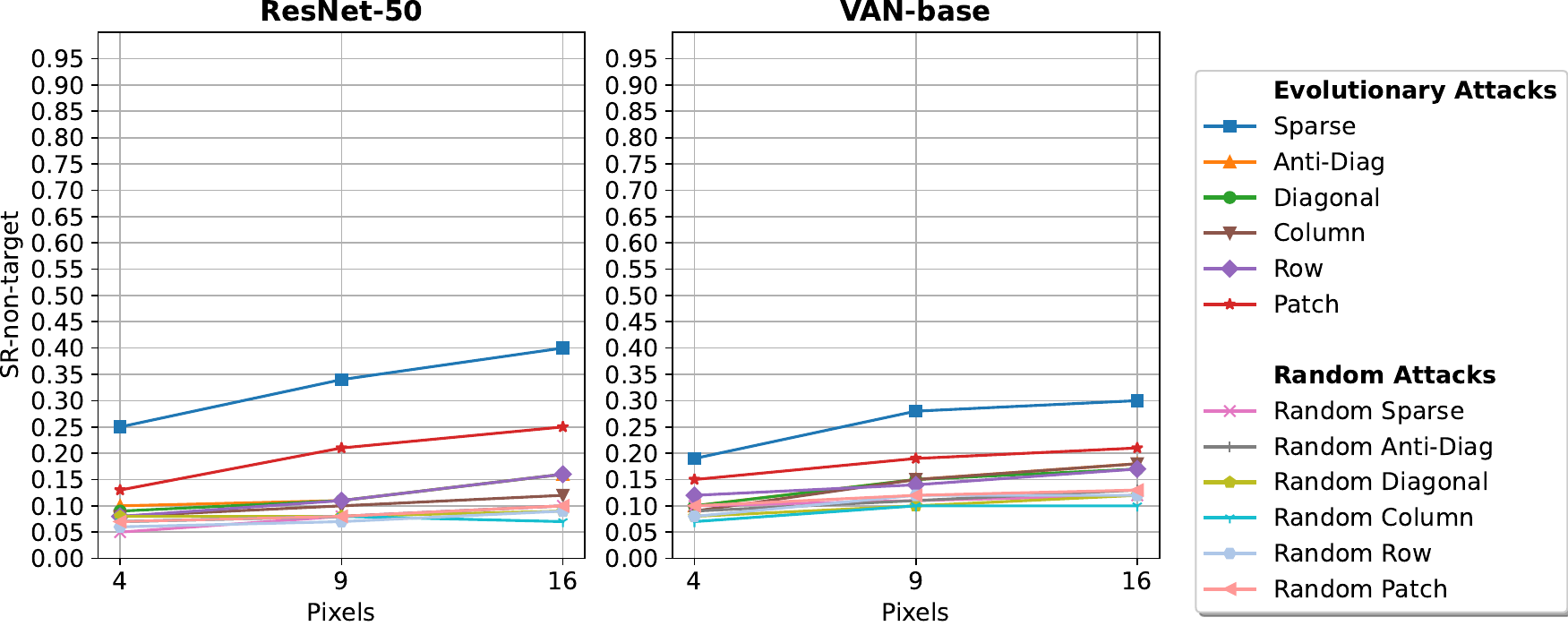}
    \caption{Untargeted Attacks}
    \label{fig:untar}
\end{figure*}
We evaluate the performances of attacks based on the SR metric and against the baselines described in Section~\ref{sec:empiric}. For instance, Figure~\ref{fig:tar} presents the SR for the targeted attacks for each selected model and the shape of the attacks, depending on the number of pixels perturbed. At first glance, we observe that for the ALIGN model, the most successful attack is the \emph{Patch Attack}, while for the others, it is the \emph{Sparse Attack}. Furthermore, the best attacks against the multimodal models achieve an SR higher than 0.35, while for the DNNs, the SR is generally lower except for ResNet-50 with the \emph{Sparse Attack} and perturbing 16 pixels. Overall, the progression of SR values with increasing perturbed pixels for the DE attacks is steeper than for the random baseline attacks. Thus demonstrating that the SR of the proposed attacks is a function of the number of perturbed pixels with an increased number of pixels yielding higher SR.

For ALIGN, Figure~\ref{fig:tar} shows that the \emph{Patch Attack} performs the best (0.74), but the value against the \emph{Sparse Attack} for 4 pixels is relatively close (0.58 for \emph{Patch} and 0.57 for \emph{Sparse}). However, the transition from 4 to 6 pixels is steeper for the \emph{Patch Attack}, starting from 0.57 and reaching 0.71, while for the \emph{Sparse Attack}, it starts from 0.56 and reaches 0.64  for the 9 pixels. Nevertheless, for the transition from 9 to 16 pixels, the slopes for both attacks seem similar. Moreover, all the \emph{Evolutionary Attacks} manage to beat the \emph{Random Attacks}, with one exception: \emph{Random Patch Attack}, which has an SR close to the ones of the \emph{Contiguous Attacks}. Besides this one, the rest of the \emph{Random} ones only achieve an SR below 40\%, even after increasing the number of pixels.

Figure~\ref{fig:tar} highlights that for AltCLIP, the \emph{Sparse Attack} is the best attack for 4 and 9 pixels with SR of 0.35 and 0.46, respectively. For 16 pixels, the most successful attack is the \emph{Patch Attack}, with the SR almost 0.5. All the \emph{Contiguous attacks} outperform the \emph{Random} ones, with the gap between those becoming more prominent with more pixels. Initially, the difference is negligible, while for 16 pixels, the gap between the \emph{Row Attack} (worst \emph{Contiguous Attack}) and \emph{Random Patch} (the best \emph{Random Attack}) is approximately 0.7. However, \emph{Random Patch} is still the best \emph{Random attack}, usually above 3-5\% from the second-best \emph{Random Attack}. Additionally, there is a big difference between the \emph{Patch} and \emph{Sparse Attacks} and the remaining \emph{Contiguous Attacks} (10-25\%).

Based on Figure~\ref{fig:tar}, CLIP is more robust to our attacks than ALIGN but less robust when compared against the other multimodal models (GroupViT, AltCLIP). The best attack remains the \emph{Sparse Attack} with values from 0.5 for 4 pixels to 0.68 for 16 pixels, followed by the \emph{Patch Attack} with values from 0.33 to 0.50. The third best attack is \emph{Row} one, and for the 16 pixels, it is followed behind with 2\% by \emph{Column} one. 

Similar to CLIP, Figure~\ref{fig:tar} shows that for GroupViT, the top two best attacks are the \emph{Sparse Attack}, reaching an SR of more than 55\% for all pixels, and the \emph{Patch Attack}, achieving a maximum SR of 47\%. Nonetheless, there is a gap between those two attacks and the others: for 16 pixels, the \emph{Row Attack} reaches a value of 0.45 and below, with 0.02 less than the \emph{Random Sparse Attack}. The best \emph{Random Attack} is the \emph{Sparse} one, which outperforms the \emph{Patch} with approximately 10\% for 16 pixels. Similarly, all the other \emph{Contiguous Attacks} are better than the other \emph{Random} ones, except for \emph{Random~Sparse}.

From Figure~\ref{fig:tar} we observe that both DNNs have similar behavior: they are more robust than any multimodal model. With ResNet-50, the SR value does not exceed 0.45 for the best attack (\emph{Sparse Attack}), while for the VAN-base, the best one stops at 0.25. Moreover, the trend regarding the best two attacks remains the same as for CLIP and GroupViT, with a significant difference regarding the gaps between those two attacks and the rest---the discrepancy is much smaller (approx. 5\% for ResNet-50 and less than 2\% for the VAN-base). Moreover, the \emph{Random Attacks} behave in the same manner as the rest of the \emph{Contiguous Attacks} with values fluctuating within the limit of 5\% for both models.

Results for the untargeted attacks are presented in Figure~\ref{fig:untar}. Overall, the trends remain the same as the targeted attacks, but the SR value is increased by approximately 25\% for ALIGN, 15\% for AltCLIP, and 10\% for CLIP and GroupViT. Interestingly, a much smaller increase is observable for both DNNs (approximately 5\%). ALIGN appears to have the worst security against the \emph{Patch Attack} for a more significant number of pixels since, using 16 pixels, we obtain a result of 99\% SR. Moreover, the \emph{Random Patch Attack} is in the third position, with an SR close to the \emph{Sparse Attack} for 16 and achieving a score of 0.85. For AltCLIP, the difference between the \emph{Random Attacks} and the \emph{Evolutionary} ones increases proportionally with the number of pixels. In addition, similar to the targeted attack, the \emph{Patch Attack} is slightly better than the \emph{Sparse} one for the 16 pixels, while for 4 and 9, the winning one is the \emph{Sparse Attack}. For CLIP and GroupViTs, the plots for ranking the effectiveness of the attacks remain the same as for targeted attacks, as is the case for the DNNs.

%% file: content/discussion.tex
At first glance, based on the results presented in Section~\ref{sec:results}, we can see that the targeted attack is more challenging than the untargeted one, which is indeed expected. However, the average difference between those two attacks for multimodal models differs from the DNNs. Also, the DNNs are more robust to our attacks than any multimodal models. An important observation is that for all setups we have at least 2-3 attacks which are better than any of the~baselines. 

Our results suggest that multimodal models with ViT (Visual Transformer)~\cite{dosovitskiy2020image} as an Image Encoder are more prone to the \emph{Sparse} than the \emph{Patch Attack}. To provide a potential explanation, we describe what happens when an image is fed to a ViT. Initially, ViT split the images into visual patches and then considers those as tokens. After this stage, those tokens are processed similarly to a normal Transformer~\cite{vaswani2017attention}. Suppose the adversary perturbs pixels on different visual patches (as you have in \emph{Sparse Attack}), then it can alter more tokens. Hence, in the self-attention layer~\cite{vaswani2017attention} (core component of a Transformer that also provides good results in terms of robustness), there is a high possibility of breaking the semantical meaning information from different visual patches. However, using any \emph{Contiguous Attack}, the adversary cannot perturb as many visual patches as would be possible using the \emph{Sparse Attack}. Additionally, the \emph{Patch Attack} outperforms any \emph{Contiguous Attack} because, with this shape, we can alter more information in a single visual patch than other contiguous pixel distributions. 

We also question why CNN-based Image Encoders (ALIGN model) are more prone to \emph{Patch Attacks}. Based on the idea from the previous paragraph, if one piece of information in the kernel is altered for those types of architectures, it is a complete disaster for the upcoming max-pooling layers. The altered information propagates to deeper layers, which try to extract correct information that can be used as features for classification.

Based on our results, ALIGN~\cite{jia2021scaling} is the most vulnerable model to the implemented $L_0$-norm perturbations. The best method to attack the ALIGN model is by using the \emph{Patch Attack} and not the \emph{Sparse} one, as it is for most of the cases in the other models. A potential explanation for this observation could be because ALIGN uses the EfficientNet~\cite{tan2021efficientnetv2} (CNN-based architecture that uses 3x3 Convolutional Layers) as an Image Encoder. This is why we have a steep increase in the SR for \emph{Patch Attack} from 4 to 9 pixels, from a patch of 2x2 to 3x3 (representing the same size as the Convolutional Layer). However, the transition from 9 to 16 pixels is smoother because we already manage to create a patch with the same dimension as the Convolutional layer, so the attack does not increase a lot in effectivness from the previous size (it increased only by 4\%, while before we have an increasing of approximate 10\%). Based on this information, the adversary can deduce some information regarding the architecture used (the size of the kernel in initial Convolutional layers) in the model and therefore here we could have a potential privacy issue. Moreover, this vulnerability of the ALIGN model against the \emph{Patch Attack} can have serious consequences since, for the untargeted attack, a patch of 16 pixels obtains a 99\% success rate. We see that the shape of the attack (\emph{Patch}) additionally confirms this vulnerability since the \emph{Random Patch Attacks} achieve high scores in both cases, especially in the untargeted setting, being the third most powerful attack.

We observe that AltCLIP is more robust than the original implementation of CLIP. This could be explained because of the architecture used in the Image Encoder. More precisely, AltCLIP~\cite{chen2022altclip} uses the ViT-L14, which has the \emph{dimension of the embedding} equals to 768, 24 \emph{layers}, \emph{width} equals to 1024, and 16 \emph{heads}, while CLIP~\cite{radford2021learning} uses the ViT-B/32 which has \emph{dimension of the embedding} equals to 512, 12 \emph{layers}, \emph{width} equals to 768, and 16 \emph{heads}. Hence, since the embeddings from AltCLIP are larger, they capture more information, and therefore, it is harder to fool those systems with the $L_{0}$-norm perturbations. Moreover, since the AltCLIP Image Encoder is deeper (based on the number of layers), it can capture more complex patterns and hierarchical features. It may exhibit better robustness and generalization---this is also proved by the Table~\ref{tab:performance_comparison}, where AltCLIP obtained the best performance compared to the other multimodal models. However, it seems that there exists a connection between the depth and the width configuration of the model with larger patches since for both non-targeted and targeted attacks in the case of 16 pixels, the \emph{Patch Attack} is slightly better, and there exists a stagnation in the \emph{Sparse Attack}. Furthermore, AltCLIP uses the XLM-R~\cite{conneau2019unsupervised} as a Text Encoder, while CLIP uses the masked self-attention transformer~\cite{dosovitskiy2020image}. Although this aspect might influence the robustness in favor of AltCLIP, the core component that improves the security aspect is the Image Encoder since we target it with our adversarial examples and do not give adversarial textual input.

For the second most robust multimodal model, GroupViT~\cite{xu2022groupvit}, we consider that the good results in terms of robustness are determined mainly by the grouping mechanism used in the Image Encoder. This new architecture uses a hierarchical group ViT---in short, besides the classical ViT approach, where the image is initially split into visual patches and projected into tokens (using Linear Projection), it also contains the Group Learnable Tokens. Those group tokens describe a different semantic notion; thus, the model also groups more semantic concepts. Since those tokens are learned during the training phase, during the grouping semantic phase, the model tries to avoid perturbed visual tokens; therefore, this mechanism increases the robustness of the model.

In addition, we observe significant differences in the SR between the DNNs and the rest of the multimodal models. A potential cause for this behavior could be related to the fact that we evaluate the DNNs on the images with specific class labels for which they were originally trained. Consequently, these DNNs contain class-specific features extracted during training on ImageNet~\cite{krizhevsky2012imagenet} which requires an attacker to perturb more pixels. In contrast to DNNs, multimodal models were trained on large corpuses of image and text data without explicit definition of target classes during training (contrastive learning). Hence, they can be used on any dataset other than the ones they were trained on. However, we argue that this flexibility comes with an increased vulnerability to pixel perturbations, which is evident from our experiments.

Lastly, a potential explanation why VAN-base~\cite{guo2023visual} is the most resilient model is based on its use of Dilated Convolutional Layers~\cite{yu2015multi}. Attacks based on contiguous pixels may be thwarted by expanding the filter's receptive field when it is widened through introducing gaps between successive elements---the main feature of dilated convolutions. As a result, the \emph{Sparse Attack} appears to be the most effective attack on VAN-base, while the \emph{Contiguous} ones behave similarly to \emph{Random Attacks}. Moreover, the robustness of the VAN-base model surpasses the ResNet-50~\cite{he2016deep}, which employs regular convolutional layers, providing further evidence for the robustness of dilated convolutions.

%% file: content/conclusion.tex
Our work analyzes the robustness of four multimodal models and two state-of-the-art unimodal DNNs against Sparse and Contiguous adversarial examples defined by the $L_0$-norm in a black-box attack scenario. For both types of attack (targeted and untargeted) and all models, we have at least two evolutionary attacks that are more effective than the rest---all multimodal models that use a ViT as an Image Encoder are most vulnerable to the \emph{Sparse Attack}. In contrast, in the CNN-based multimodal model (ALIGN), the most effective attack is represented by a contiguous pixel perturbation in the form of a \emph{Patch}. This attack prove to be so powerful against ALIGN that with an increased patch dimension of 4x4 (less than \textbf{0.02}\% of the image area), in a non-targeted scenario, we achieved a \textbf{99}\% SR. Furthermore, based on the overall SR, we rank the most secure models as follows: VAN-base, ResNet-50, AltCLIP, GroupViT, CLIP-B/32, and ALIGN.
Our results also point towards different characteristics of AI model architectures (e.g., tokenization, convolutions, dilated convolutions) that may be responsible for different levels of robustness against our attacks. When contrasting the results for multimodal and unimodal models, we observe that there may be an essential trade-off between the robustness of the model and its adaptability to be used in diverse tasks (i.e., zero-shot capability). Both of these aspects should be further investigated in a follow-up study.

For future work, we propose to study the attacks on the $L_0$-norm for other types of multimodal models that focus on object detection and text generation, such as IDEFICS~\cite{laurenccon2023obelisc}, Kosmos-2~\cite{peng2023kosmos}, and Gemini~\cite{team2023gemini}. Moreover, we want to do more investigation on the contiguous attacks and focus on finding the best hyperparameters for the evolution algorithm used for those specific patterns, such that we can construct a systematic search setup, where we can infer different information regarding the model architecture from the observed attack performance.

%% file: content/appendix.tex
\section{Exponential Strategy in Differential Evolution}
\label{sec:exp_algo}

Algorithm~\ref{alg:exp_strat} was used for the Crossover operation - Exponential Strategy in the Differential Evolution. The idea of this algorithm is to recombine the agents based on some $crossover_{rate}$---how much information from the old agents ($agent_{old}$) should also be preserved in the new ones ($agent_{cand}$). Thus, we directly modify the new agents in the $agent_{cand}$ vector and return it for further processing. This code was adapted from the \texttt{Differential Evolution} implemented in \texttt{Scipy} ~\cite{scipy_doc}.

\begin{algorithm}[htb]
\caption{Exponential Strategy~\cite{scipy_doc}}
\label{alg:exp_strat}
\begin{algorithmic}[1]
\Function{Crossover Exp}{$agent_{old}, agent_{cand}, crossover_{rate}$}
    \State $big_r \gets \text{random index value of the } agent_{cand}$
    \State $i \gets 0$
    \While{$i < \text{size of }agent_{cand}$}
        \State $r \gets$ \text{random number between 0 and 1}
        \If{$crossover_{rate} < r$}
            \State \textbf{break}
        \EndIf
        \State $agent_{cand}[big_r] \gets agent_{old}[big_r]$
        \State $big_r \gets (big_r + 1) \mod agent_{cand}\_size$
        \State $i \gets i + 1$
    \EndWhile
    \State \textbf{return} $agent_{cand}$
\EndFunction
\end{algorithmic}
\end{algorithm}

%% file: main.bbl
\begin{thebibliography}{65}


\ifx \showCODEN    \undefined \def \showCODEN     #1{\unskip}     \fi
\ifx \showDOI      \undefined \def \showDOI       #1{#1}\fi
\ifx \showISBNx    \undefined \def \showISBNx     #1{\unskip}     \fi
\ifx \showISBNxiii \undefined \def \showISBNxiii  #1{\unskip}     \fi
\ifx \showISSN     \undefined \def \showISSN      #1{\unskip}     \fi
\ifx \showLCCN     \undefined \def \showLCCN      #1{\unskip}     \fi
\ifx \shownote     \undefined \def \shownote      #1{#1}          \fi
\ifx \showarticletitle \undefined \def \showarticletitle #1{#1}   \fi
\ifx \showURL      \undefined \def \showURL       {\relax}        \fi
\providecommand\bibfield[2]{#2}
\providecommand\bibinfo[2]{#2}
\providecommand\natexlab[1]{#1}
\providecommand\showeprint[2][]{arXiv:#2}

\bibitem[Alrasheedi and Zhong(2023)]%
        {alrasheedi2023imperceptible}
\bibfield{author}{\bibinfo{person}{Fahad Alrasheedi} {and} \bibinfo{person}{Xin Zhong}.} \bibinfo{year}{2023}\natexlab{}.
\newblock \showarticletitle{Imperceptible Adversarial Attack on Deep Neural Networks from Image Boundary}.
\newblock \bibinfo{journal}{\emph{arXiv preprint arXiv:2308.15344}} (\bibinfo{year}{2023}).
\newblock


\bibitem[Athalye et~al\mbox{.}(2018)]%
        {athalye2018obfuscated}
\bibfield{author}{\bibinfo{person}{Anish Athalye}, \bibinfo{person}{Nicholas Carlini}, {and} \bibinfo{person}{David Wagner}.} \bibinfo{year}{2018}\natexlab{}.
\newblock \showarticletitle{Obfuscated gradients give a false sense of security: Circumventing defenses to adversarial examples}. In \bibinfo{booktitle}{\emph{International conference on machine learning}}. PMLR, \bibinfo{pages}{274--283}.
\newblock


\bibitem[Awadalla et~al\mbox{.}(2023)]%
        {awadalla2023openflamingo}
\bibfield{author}{\bibinfo{person}{Anas Awadalla}, \bibinfo{person}{Irena Gao}, \bibinfo{person}{Josh Gardner}, \bibinfo{person}{Jack Hessel}, \bibinfo{person}{Yusuf Hanafy}, \bibinfo{person}{Wanrong Zhu}, \bibinfo{person}{Kalyani Marathe}, \bibinfo{person}{Yonatan Bitton}, \bibinfo{person}{Samir Gadre}, \bibinfo{person}{Shiori Sagawa}, {et~al\mbox{.}}} \bibinfo{year}{2023}\natexlab{}.
\newblock \showarticletitle{Openflamingo: An open-source framework for training large autoregressive vision-language models}.
\newblock \bibinfo{journal}{\emph{arXiv preprint arXiv:2308.01390}} (\bibinfo{year}{2023}).
\newblock


\bibitem[Bai et~al\mbox{.}(2021)]%
        {bai2021inconspicuous}
\bibfield{author}{\bibinfo{person}{Tao Bai}, \bibinfo{person}{Jinqi Luo}, {and} \bibinfo{person}{Jun Zhao}.} \bibinfo{year}{2021}\natexlab{}.
\newblock \showarticletitle{Inconspicuous adversarial patches for fooling image-recognition systems on mobile devices}.
\newblock \bibinfo{journal}{\emph{IEEE Internet of Things Journal}} \bibinfo{volume}{9}, \bibinfo{number}{12} (\bibinfo{year}{2021}), \bibinfo{pages}{9515--9524}.
\newblock


\bibitem[Brendel et~al\mbox{.}(2017)]%
        {brendel2017decision}
\bibfield{author}{\bibinfo{person}{Wieland Brendel}, \bibinfo{person}{Jonas Rauber}, {and} \bibinfo{person}{Matthias Bethge}.} \bibinfo{year}{2017}\natexlab{}.
\newblock \showarticletitle{Decision-based adversarial attacks: Reliable attacks against black-box machine learning models}.
\newblock \bibinfo{journal}{\emph{arXiv preprint arXiv:1712.04248}} (\bibinfo{year}{2017}).
\newblock


\bibitem[Brown et~al\mbox{.}(2017)]%
        {brown2017adversarial}
\bibfield{author}{\bibinfo{person}{Tom~B Brown}, \bibinfo{person}{Dandelion Man{\'e}}, \bibinfo{person}{Aurko Roy}, \bibinfo{person}{Mart{\'\i}n Abadi}, {and} \bibinfo{person}{Justin Gilmer}.} \bibinfo{year}{2017}\natexlab{}.
\newblock \showarticletitle{Adversarial patch}.
\newblock \bibinfo{journal}{\emph{arXiv preprint arXiv:1712.09665}} (\bibinfo{year}{2017}).
\newblock


\bibitem[Cao et~al\mbox{.}(2023)]%
        {cao2023less}
\bibfield{author}{\bibinfo{person}{Liangliang Cao}, \bibinfo{person}{Bowen Zhang}, \bibinfo{person}{Chen Chen}, \bibinfo{person}{Yinfei Yang}, \bibinfo{person}{Xianzhi Du}, \bibinfo{person}{Wencong Zhang}, \bibinfo{person}{Zhiyun Lu}, {and} \bibinfo{person}{Yantao Zheng}.} \bibinfo{year}{2023}\natexlab{}.
\newblock \showarticletitle{Less is More: Removing Text-regions Improves CLIP Training Efficiency and Robustness}.
\newblock \bibinfo{journal}{\emph{arXiv preprint arXiv:2305.05095}} (\bibinfo{year}{2023}).
\newblock


\bibitem[Carlini and Wagner(2017)]%
        {carlini2017towards}
\bibfield{author}{\bibinfo{person}{Nicholas Carlini} {and} \bibinfo{person}{David Wagner}.} \bibinfo{year}{2017}\natexlab{}.
\newblock \showarticletitle{Towards evaluating the robustness of neural networks}. In \bibinfo{booktitle}{\emph{2017 ieee symposium on security and privacy (sp)}}. Ieee, \bibinfo{pages}{39--57}.
\newblock


\bibitem[Chen et~al\mbox{.}(2017)]%
        {chen2017zoo}
\bibfield{author}{\bibinfo{person}{Pin-Yu Chen}, \bibinfo{person}{Huan Zhang}, \bibinfo{person}{Yash Sharma}, \bibinfo{person}{Jinfeng Yi}, {and} \bibinfo{person}{Cho-Jui Hsieh}.} \bibinfo{year}{2017}\natexlab{}.
\newblock \showarticletitle{Zoo: Zeroth order optimization based black-box attacks to deep neural networks without training substitute models}. In \bibinfo{booktitle}{\emph{Proceedings of the 10th ACM workshop on artificial intelligence and security}}. \bibinfo{pages}{15--26}.
\newblock


\bibitem[Chen et~al\mbox{.}(2022)]%
        {chen2022altclip}
\bibfield{author}{\bibinfo{person}{Zhongzhi Chen}, \bibinfo{person}{Guang Liu}, \bibinfo{person}{Bo-Wen Zhang}, \bibinfo{person}{Fulong Ye}, \bibinfo{person}{Qinghong Yang}, {and} \bibinfo{person}{Ledell Wu}.} \bibinfo{year}{2022}\natexlab{}.
\newblock \showarticletitle{Altclip: Altering the language encoder in clip for extended language capabilities}.
\newblock \bibinfo{journal}{\emph{arXiv preprint arXiv:2211.06679}} (\bibinfo{year}{2022}).
\newblock


\bibitem[Chindaudom et~al\mbox{.}(2020)]%
        {chindaudom2020adversarialqr}
\bibfield{author}{\bibinfo{person}{Aran Chindaudom}, \bibinfo{person}{Prarinya Siritanawan}, \bibinfo{person}{Karin Sumongkayothin}, {and} \bibinfo{person}{Kazunori Kotani}.} \bibinfo{year}{2020}\natexlab{}.
\newblock \showarticletitle{AdversarialQR: An adversarial patch in QR code format}. In \bibinfo{booktitle}{\emph{2020 Joint 9th International Conference on Informatics, Electronics \& Vision (ICIEV) and 2020 4th International Conference on Imaging, Vision \& Pattern Recognition (icIVPR)}}. IEEE, \bibinfo{pages}{1--6}.
\newblock


\bibitem[Chindaudom et~al\mbox{.}(2022)]%
        {chindaudom2022surreptitious}
\bibfield{author}{\bibinfo{person}{Aran Chindaudom}, \bibinfo{person}{Prarinya Siritanawan}, \bibinfo{person}{Karin Sumongkayothin}, {and} \bibinfo{person}{Kazunori Kotani}.} \bibinfo{year}{2022}\natexlab{}.
\newblock \showarticletitle{Surreptitious Adversarial Examples through Functioning QR Code}.
\newblock \bibinfo{journal}{\emph{Journal of Imaging}} \bibinfo{volume}{8}, \bibinfo{number}{5} (\bibinfo{year}{2022}), \bibinfo{pages}{122}.
\newblock


\bibitem[Conneau et~al\mbox{.}(2019)]%
        {conneau2019unsupervised}
\bibfield{author}{\bibinfo{person}{Alexis Conneau}, \bibinfo{person}{Kartikay Khandelwal}, \bibinfo{person}{Naman Goyal}, \bibinfo{person}{Vishrav Chaudhary}, \bibinfo{person}{Guillaume Wenzek}, \bibinfo{person}{Francisco Guzm{\'a}n}, \bibinfo{person}{Edouard Grave}, \bibinfo{person}{Myle Ott}, \bibinfo{person}{Luke Zettlemoyer}, {and} \bibinfo{person}{Veselin Stoyanov}.} \bibinfo{year}{2019}\natexlab{}.
\newblock \showarticletitle{Unsupervised cross-lingual representation learning at scale}.
\newblock \bibinfo{journal}{\emph{arXiv preprint arXiv:1911.02116}} (\bibinfo{year}{2019}).
\newblock


\bibitem[Demir and Unal(2018)]%
        {demir2018patch}
\bibfield{author}{\bibinfo{person}{Ugur Demir} {and} \bibinfo{person}{Gozde Unal}.} \bibinfo{year}{2018}\natexlab{}.
\newblock \showarticletitle{Patch-based image inpainting with generative adversarial networks}.
\newblock \bibinfo{journal}{\emph{arXiv preprint arXiv:1803.07422}} (\bibinfo{year}{2018}).
\newblock


\bibitem[Dosovitskiy et~al\mbox{.}(2020)]%
        {dosovitskiy2020image}
\bibfield{author}{\bibinfo{person}{Alexey Dosovitskiy}, \bibinfo{person}{Lucas Beyer}, \bibinfo{person}{Alexander Kolesnikov}, \bibinfo{person}{Dirk Weissenborn}, \bibinfo{person}{Xiaohua Zhai}, \bibinfo{person}{Thomas Unterthiner}, \bibinfo{person}{Mostafa Dehghani}, \bibinfo{person}{Matthias Minderer}, \bibinfo{person}{Georg Heigold}, \bibinfo{person}{Sylvain Gelly}, {et~al\mbox{.}}} \bibinfo{year}{2020}\natexlab{}.
\newblock \showarticletitle{An image is worth 16x16 words: Transformers for image recognition at scale}.
\newblock \bibinfo{journal}{\emph{arXiv preprint arXiv:2010.11929}} (\bibinfo{year}{2020}).
\newblock


\bibitem[Fort(2021)]%
        {fort2021pixels}
\bibfield{author}{\bibinfo{person}{Stanislav Fort}.} \bibinfo{year}{2021}\natexlab{}.
\newblock \showarticletitle{Pixels still beat text: attacking the OpenAI CLIP model with text patches and adversarial pixel perturbations}.
\newblock \bibinfo{journal}{\emph{Stanislav Fort [Internet]}}  \bibinfo{volume}{5} (\bibinfo{year}{2021}).
\newblock


\bibitem[Freiberger et~al\mbox{.}(2023)]%
        {freiberger2023clipmasterprints}
\bibfield{author}{\bibinfo{person}{Matthias Freiberger}, \bibinfo{person}{Peter Kun}, \bibinfo{person}{Anders~Sundnes L{\o}vlie}, {and} \bibinfo{person}{Sebastian Risi}.} \bibinfo{year}{2023}\natexlab{}.
\newblock \showarticletitle{CLIPMasterPrints: Fooling Contrastive Language-Image Pre-training Using Latent Variable Evolution}.
\newblock \bibinfo{journal}{\emph{arXiv preprint arXiv:2307.03798}} (\bibinfo{year}{2023}).
\newblock


\bibitem[Ghosh et~al\mbox{.}(2022)]%
        {ghosh2022black}
\bibfield{author}{\bibinfo{person}{Arka Ghosh}, \bibinfo{person}{Sankha~Subhra Mullick}, \bibinfo{person}{Shounak Datta}, \bibinfo{person}{Swagatam Das}, \bibinfo{person}{Asit~Kr Das}, {and} \bibinfo{person}{Rammohan Mallipeddi}.} \bibinfo{year}{2022}\natexlab{}.
\newblock \showarticletitle{A black-box adversarial attack strategy with adjustable sparsity and generalizability for deep image classifiers}.
\newblock \bibinfo{journal}{\emph{Pattern Recognition}}  \bibinfo{volume}{122} (\bibinfo{year}{2022}), \bibinfo{pages}{108279}.
\newblock


\bibitem[Goodfellow et~al\mbox{.}(2014)]%
        {goodfellow2014explaining}
\bibfield{author}{\bibinfo{person}{Ian~J Goodfellow}, \bibinfo{person}{Jonathon Shlens}, {and} \bibinfo{person}{Christian Szegedy}.} \bibinfo{year}{2014}\natexlab{}.
\newblock \showarticletitle{Explaining and harnessing adversarial examples}.
\newblock \bibinfo{journal}{\emph{arXiv preprint arXiv:1412.6572}} (\bibinfo{year}{2014}).
\newblock


\bibitem[Guo et~al\mbox{.}(2023)]%
        {guo2023visual}
\bibfield{author}{\bibinfo{person}{Meng-Hao Guo}, \bibinfo{person}{Cheng-Ze Lu}, \bibinfo{person}{Zheng-Ning Liu}, \bibinfo{person}{Ming-Ming Cheng}, {and} \bibinfo{person}{Shi-Min Hu}.} \bibinfo{year}{2023}\natexlab{}.
\newblock \showarticletitle{Visual attention network}.
\newblock \bibinfo{journal}{\emph{Computational Visual Media}} \bibinfo{volume}{9}, \bibinfo{number}{4} (\bibinfo{year}{2023}), \bibinfo{pages}{733--752}.
\newblock


\bibitem[Haas et~al\mbox{.}(2023)]%
        {haas2023learning}
\bibfield{author}{\bibinfo{person}{Lukas Haas}, \bibinfo{person}{Silas Alberti}, {and} \bibinfo{person}{Michal Skreta}.} \bibinfo{year}{2023}\natexlab{}.
\newblock \bibinfo{title}{Learning Generalized Zero-Shot Learners for Open-Domain Image Geolocalization}.
\newblock
\newblock
\showeprint[arxiv]{2302.00275}~[cs.CV]


\bibitem[He et~al\mbox{.}(2016)]%
        {he2016deep}
\bibfield{author}{\bibinfo{person}{Kaiming He}, \bibinfo{person}{Xiangyu Zhang}, \bibinfo{person}{Shaoqing Ren}, {and} \bibinfo{person}{Jian Sun}.} \bibinfo{year}{2016}\natexlab{}.
\newblock \showarticletitle{Deep residual learning for image recognition}. In \bibinfo{booktitle}{\emph{Proceedings of the IEEE conference on computer vision and pattern recognition}}. \bibinfo{pages}{770--778}.
\newblock


\bibitem[Hu et~al\mbox{.}(2021b)]%
        {hu2021advhash}
\bibfield{author}{\bibinfo{person}{Shengshan Hu}, \bibinfo{person}{Yechao Zhang}, \bibinfo{person}{Xiaogeng Liu}, \bibinfo{person}{Leo~Yu Zhang}, \bibinfo{person}{Minghui Li}, {and} \bibinfo{person}{Hai Jin}.} \bibinfo{year}{2021}\natexlab{b}.
\newblock \showarticletitle{Advhash: Set-to-set targeted attack on deep hashing with one single adversarial patch}. In \bibinfo{booktitle}{\emph{Proceedings of the 29th ACM International Conference on Multimedia}}. \bibinfo{pages}{2335--2343}.
\newblock


\bibitem[Hu et~al\mbox{.}(2021a)]%
        {hu2021naturalistic}
\bibfield{author}{\bibinfo{person}{Yu-Chih-Tuan Hu}, \bibinfo{person}{Bo-Han Kung}, \bibinfo{person}{Daniel~Stanley Tan}, \bibinfo{person}{Jun-Cheng Chen}, \bibinfo{person}{Kai-Lung Hua}, {and} \bibinfo{person}{Wen-Huang Cheng}.} \bibinfo{year}{2021}\natexlab{a}.
\newblock \showarticletitle{Naturalistic physical adversarial patch for object detectors}. In \bibinfo{booktitle}{\emph{Proceedings of the IEEE/CVF International Conference on Computer Vision}}. \bibinfo{pages}{7848--7857}.
\newblock


\bibitem[Jere et~al\mbox{.}(2019)]%
        {jere2019scratch}
\bibfield{author}{\bibinfo{person}{Malhar Jere}, \bibinfo{person}{Loris Rossi}, \bibinfo{person}{Briland Hitaj}, \bibinfo{person}{Gabriela Ciocarlie}, \bibinfo{person}{Giacomo Boracchi}, {and} \bibinfo{person}{Farinaz Koushanfar}.} \bibinfo{year}{2019}\natexlab{}.
\newblock \showarticletitle{Scratch that! An evolution-based adversarial attack against neural networks}.
\newblock \bibinfo{journal}{\emph{arXiv preprint arXiv:1912.02316}} (\bibinfo{year}{2019}).
\newblock


\bibitem[Jia et~al\mbox{.}(2021)]%
        {jia2021scaling}
\bibfield{author}{\bibinfo{person}{Chao Jia}, \bibinfo{person}{Yinfei Yang}, \bibinfo{person}{Ye Xia}, \bibinfo{person}{Yi-Ting Chen}, \bibinfo{person}{Zarana Parekh}, \bibinfo{person}{Hieu Pham}, \bibinfo{person}{Quoc Le}, \bibinfo{person}{Yun-Hsuan Sung}, \bibinfo{person}{Zhen Li}, {and} \bibinfo{person}{Tom Duerig}.} \bibinfo{year}{2021}\natexlab{}.
\newblock \showarticletitle{Scaling up visual and vision-language representation learning with noisy text supervision}. In \bibinfo{booktitle}{\emph{International conference on machine learning}}. PMLR, \bibinfo{pages}{4904--4916}.
\newblock


\bibitem[Karmon et~al\mbox{.}(2018)]%
        {karmon2018lavan}
\bibfield{author}{\bibinfo{person}{Danny Karmon}, \bibinfo{person}{Daniel Zoran}, {and} \bibinfo{person}{Yoav Goldberg}.} \bibinfo{year}{2018}\natexlab{}.
\newblock \showarticletitle{Lavan: Localized and visible adversarial noise}. In \bibinfo{booktitle}{\emph{International Conference on Machine Learning}}. PMLR, \bibinfo{pages}{2507--2515}.
\newblock


\bibitem[Krizhevsky et~al\mbox{.}(2012)]%
        {krizhevsky2012imagenet}
\bibfield{author}{\bibinfo{person}{Alex Krizhevsky}, \bibinfo{person}{Ilya Sutskever}, {and} \bibinfo{person}{Geoffrey~E Hinton}.} \bibinfo{year}{2012}\natexlab{}.
\newblock \showarticletitle{Imagenet classification with deep convolutional neural networks}.
\newblock \bibinfo{journal}{\emph{Advances in neural information processing systems}}  \bibinfo{volume}{25} (\bibinfo{year}{2012}).
\newblock


\bibitem[Kurakin et~al\mbox{.}(2016)]%
        {kurakin2016adversarial}
\bibfield{author}{\bibinfo{person}{Alexey Kurakin}, \bibinfo{person}{Ian Goodfellow}, {and} \bibinfo{person}{Samy Bengio}.} \bibinfo{year}{2016}\natexlab{}.
\newblock \showarticletitle{Adversarial machine learning at scale}.
\newblock \bibinfo{journal}{\emph{arXiv preprint arXiv:1611.01236}} (\bibinfo{year}{2016}).
\newblock


\bibitem[Lauren{\c{c}}on et~al\mbox{.}(2023)]%
        {laurenccon2023obelisc}
\bibfield{author}{\bibinfo{person}{Hugo Lauren{\c{c}}on}, \bibinfo{person}{Lucile Saulnier}, \bibinfo{person}{L{\'e}o Tronchon}, \bibinfo{person}{Stas Bekman}, \bibinfo{person}{Amanpreet Singh}, \bibinfo{person}{Anton Lozhkov}, \bibinfo{person}{Thomas Wang}, \bibinfo{person}{Siddharth Karamcheti}, \bibinfo{person}{Alexander~M Rush}, \bibinfo{person}{Douwe Kiela}, {et~al\mbox{.}}} \bibinfo{year}{2023}\natexlab{}.
\newblock \showarticletitle{Obelisc: An open web-scale filtered dataset of interleaved image-text documents}.
\newblock \bibinfo{journal}{\emph{arXiv preprint arXiv:2306.16527}} (\bibinfo{year}{2023}).
\newblock


\bibitem[Li et~al\mbox{.}(2022)]%
        {li2022blip}
\bibfield{author}{\bibinfo{person}{Junnan Li}, \bibinfo{person}{Dongxu Li}, \bibinfo{person}{Caiming Xiong}, {and} \bibinfo{person}{Steven Hoi}.} \bibinfo{year}{2022}\natexlab{}.
\newblock \showarticletitle{Blip: Bootstrapping language-image pre-training for unified vision-language understanding and generation}. In \bibinfo{booktitle}{\emph{International Conference on Machine Learning}}. PMLR, \bibinfo{pages}{12888--12900}.
\newblock


\bibitem[Liu et~al\mbox{.}(2019)]%
        {liu2019perceptual}
\bibfield{author}{\bibinfo{person}{Aishan Liu}, \bibinfo{person}{Xianglong Liu}, \bibinfo{person}{Jiaxin Fan}, \bibinfo{person}{Yuqing Ma}, \bibinfo{person}{Anlan Zhang}, \bibinfo{person}{Huiyuan Xie}, {and} \bibinfo{person}{Dacheng Tao}.} \bibinfo{year}{2019}\natexlab{}.
\newblock \showarticletitle{Perceptual-sensitive gan for generating adversarial patches}. In \bibinfo{booktitle}{\emph{Proceedings of the AAAI conference on artificial intelligence}}, Vol.~\bibinfo{volume}{33}. \bibinfo{pages}{1028--1035}.
\newblock


\bibitem[Madry et~al\mbox{.}(2017)]%
        {madry2017towards}
\bibfield{author}{\bibinfo{person}{Aleksander Madry}, \bibinfo{person}{Aleksandar Makelov}, \bibinfo{person}{Ludwig Schmidt}, \bibinfo{person}{Dimitris Tsipras}, {and} \bibinfo{person}{Adrian Vladu}.} \bibinfo{year}{2017}\natexlab{}.
\newblock \showarticletitle{Towards deep learning models resistant to adversarial attacks}.
\newblock \bibinfo{journal}{\emph{arXiv preprint arXiv:1706.06083}} (\bibinfo{year}{2017}).
\newblock


\bibitem[Mao et~al\mbox{.}(2022)]%
        {mao2022understanding}
\bibfield{author}{\bibinfo{person}{Chengzhi Mao}, \bibinfo{person}{Scott Geng}, \bibinfo{person}{Junfeng Yang}, \bibinfo{person}{Xin Wang}, {and} \bibinfo{person}{Carl Vondrick}.} \bibinfo{year}{2022}\natexlab{}.
\newblock \showarticletitle{Understanding zero-shot adversarial robustness for large-scale models}.
\newblock \bibinfo{journal}{\emph{arXiv preprint arXiv:2212.07016}} (\bibinfo{year}{2022}).
\newblock


\bibitem[Moosavi-Dezfooli et~al\mbox{.}(2016)]%
        {moosavi2016deepfool}
\bibfield{author}{\bibinfo{person}{Seyed-Mohsen Moosavi-Dezfooli}, \bibinfo{person}{Alhussein Fawzi}, {and} \bibinfo{person}{Pascal Frossard}.} \bibinfo{year}{2016}\natexlab{}.
\newblock \showarticletitle{Deepfool: a simple and accurate method to fool deep neural networks}. In \bibinfo{booktitle}{\emph{Proceedings of the IEEE conference on computer vision and pattern recognition}}. \bibinfo{pages}{2574--2582}.
\newblock


\bibitem[Nam and Kil(2023)]%
        {nam2023aesop}
\bibfield{author}{\bibinfo{person}{Wonhong Nam} {and} \bibinfo{person}{Hyunyoung Kil}.} \bibinfo{year}{2023}\natexlab{}.
\newblock \showarticletitle{AESOP: Adjustable Exhaustive Search for One-Pixel Attacks in Deep Neural Networks}.
\newblock \bibinfo{journal}{\emph{Applied Sciences}} \bibinfo{volume}{13}, \bibinfo{number}{8} (\bibinfo{year}{2023}), \bibinfo{pages}{5092}.
\newblock


\bibitem[Papernot et~al\mbox{.}(2016)]%
        {papernot2016limitations}
\bibfield{author}{\bibinfo{person}{Nicolas Papernot}, \bibinfo{person}{Patrick McDaniel}, \bibinfo{person}{Somesh Jha}, \bibinfo{person}{Matt Fredrikson}, \bibinfo{person}{Z~Berkay Celik}, {and} \bibinfo{person}{Ananthram Swami}.} \bibinfo{year}{2016}\natexlab{}.
\newblock \showarticletitle{The limitations of deep learning in adversarial settings}. In \bibinfo{booktitle}{\emph{2016 IEEE European symposium on security and privacy (EuroS\&P)}}. IEEE, \bibinfo{pages}{372--387}.
\newblock


\bibitem[Pavlitskaya et~al\mbox{.}(2022)]%
        {pavlitskaya2022feasibility}
\bibfield{author}{\bibinfo{person}{Svetlana Pavlitskaya}, \bibinfo{person}{Bianca-Marina Cod{\u{a}}u}, {and} \bibinfo{person}{J~Marius Z{\"o}llner}.} \bibinfo{year}{2022}\natexlab{}.
\newblock \showarticletitle{Feasibility of inconspicuous GAN-generated adversarial patches against object detection}.
\newblock \bibinfo{journal}{\emph{arXiv preprint arXiv:2207.07347}} (\bibinfo{year}{2022}).
\newblock


\bibitem[Peng et~al\mbox{.}(2023)]%
        {peng2023kosmos}
\bibfield{author}{\bibinfo{person}{Zhiliang Peng}, \bibinfo{person}{Wenhui Wang}, \bibinfo{person}{Li Dong}, \bibinfo{person}{Yaru Hao}, \bibinfo{person}{Shaohan Huang}, \bibinfo{person}{Shuming Ma}, {and} \bibinfo{person}{Furu Wei}.} \bibinfo{year}{2023}\natexlab{}.
\newblock \showarticletitle{Kosmos-2: Grounding Multimodal Large Language Models to the World}.
\newblock \bibinfo{journal}{\emph{arXiv preprint arXiv:2306.14824}} (\bibinfo{year}{2023}).
\newblock


\bibitem[Pham et~al\mbox{.}(2023)]%
        {pham2023combined}
\bibfield{author}{\bibinfo{person}{Hieu Pham}, \bibinfo{person}{Zihang Dai}, \bibinfo{person}{Golnaz Ghiasi}, \bibinfo{person}{Kenji Kawaguchi}, \bibinfo{person}{Hanxiao Liu}, \bibinfo{person}{Adams~Wei Yu}, \bibinfo{person}{Jiahui Yu}, \bibinfo{person}{Yi-Ting Chen}, \bibinfo{person}{Minh-Thang Luong}, \bibinfo{person}{Yonghui Wu}, {et~al\mbox{.}}} \bibinfo{year}{2023}\natexlab{}.
\newblock \showarticletitle{Combined scaling for zero-shot transfer learning}.
\newblock \bibinfo{journal}{\emph{Neurocomputing}}  \bibinfo{volume}{555} (\bibinfo{year}{2023}), \bibinfo{pages}{126658}.
\newblock


\bibitem[Qiu et~al\mbox{.}(2021)]%
        {qiu2021black}
\bibfield{author}{\bibinfo{person}{Hao Qiu}, \bibinfo{person}{Leonardo~Lucio Custode}, {and} \bibinfo{person}{Giovanni Iacca}.} \bibinfo{year}{2021}\natexlab{}.
\newblock \showarticletitle{Black-box adversarial attacks using evolution strategies}. In \bibinfo{booktitle}{\emph{Proceedings of the Genetic and Evolutionary Computation Conference Companion}}. \bibinfo{pages}{1827--1833}.
\newblock


\bibitem[Qiu et~al\mbox{.}(2022)]%
        {qiu2022multimodal}
\bibfield{author}{\bibinfo{person}{Jielin Qiu}, \bibinfo{person}{Yi Zhu}, \bibinfo{person}{Xingjian Shi}, \bibinfo{person}{Florian Wenzel}, \bibinfo{person}{Zhiqiang Tang}, \bibinfo{person}{Ding Zhao}, \bibinfo{person}{Bo Li}, {and} \bibinfo{person}{Mu Li}.} \bibinfo{year}{2022}\natexlab{}.
\newblock \showarticletitle{Are Multimodal Models Robust to Image and Text Perturbations?}
\newblock \bibinfo{journal}{\emph{arXiv preprint arXiv:2212.08044}} (\bibinfo{year}{2022}).
\newblock


\bibitem[Radford et~al\mbox{.}(2021)]%
        {radford2021learning}
\bibfield{author}{\bibinfo{person}{Alec Radford}, \bibinfo{person}{Jong~Wook Kim}, \bibinfo{person}{Chris Hallacy}, \bibinfo{person}{Aditya Ramesh}, \bibinfo{person}{Gabriel Goh}, \bibinfo{person}{Sandhini Agarwal}, \bibinfo{person}{Girish Sastry}, \bibinfo{person}{Amanda Askell}, \bibinfo{person}{Pamela Mishkin}, \bibinfo{person}{Jack Clark}, {et~al\mbox{.}}} \bibinfo{year}{2021}\natexlab{}.
\newblock \showarticletitle{Learning transferable visual models from natural language supervision}. In \bibinfo{booktitle}{\emph{International conference on machine learning}}. PMLR, \bibinfo{pages}{8748--8763}.
\newblock


\bibitem[Schlarmann and Hein(2023)]%
        {schlarmann2023adversarial}
\bibfield{author}{\bibinfo{person}{Christian Schlarmann} {and} \bibinfo{person}{Matthias Hein}.} \bibinfo{year}{2023}\natexlab{}.
\newblock \showarticletitle{On the adversarial robustness of multi-modal foundation models}. In \bibinfo{booktitle}{\emph{Proceedings of the IEEE/CVF International Conference on Computer Vision}}. \bibinfo{pages}{3677--3685}.
\newblock


\bibitem[{SciPy Contributors}(2024)]%
        {scipy_doc}
\bibfield{author}{\bibinfo{person}{{SciPy Contributors}}.} \bibinfo{year}{Accessed: 2024}\natexlab{}.
\newblock \bibinfo{booktitle}{\emph{scipy.optimize.differential\_evolution}}.
\newblock


\bibitem[Simonyan and Zisserman(2014)]%
        {simonyan2014very}
\bibfield{author}{\bibinfo{person}{Karen Simonyan} {and} \bibinfo{person}{Andrew Zisserman}.} \bibinfo{year}{2014}\natexlab{}.
\newblock \showarticletitle{Very deep convolutional networks for large-scale image recognition}.
\newblock \bibinfo{journal}{\emph{arXiv preprint arXiv:1409.1556}} (\bibinfo{year}{2014}).
\newblock


\bibitem[Storn and Price(1997)]%
        {storn1997differential}
\bibfield{author}{\bibinfo{person}{Rainer Storn} {and} \bibinfo{person}{Kenneth Price}.} \bibinfo{year}{1997}\natexlab{}.
\newblock \showarticletitle{Differential evolution--a simple and efficient heuristic for global optimization over continuous spaces}.
\newblock \bibinfo{journal}{\emph{Journal of global optimization}}  \bibinfo{volume}{11} (\bibinfo{year}{1997}), \bibinfo{pages}{341--359}.
\newblock


\bibitem[Su et~al\mbox{.}(2019)]%
        {su2019one}
\bibfield{author}{\bibinfo{person}{Jiawei Su}, \bibinfo{person}{Danilo~Vasconcellos Vargas}, {and} \bibinfo{person}{Kouichi Sakurai}.} \bibinfo{year}{2019}\natexlab{}.
\newblock \showarticletitle{One pixel attack for fooling deep neural networks}.
\newblock \bibinfo{journal}{\emph{IEEE Transactions on Evolutionary Computation}} \bibinfo{volume}{23}, \bibinfo{number}{5} (\bibinfo{year}{2019}), \bibinfo{pages}{828--841}.
\newblock


\bibitem[Szegedy et~al\mbox{.}(2015)]%
        {szegedy2015going}
\bibfield{author}{\bibinfo{person}{Christian Szegedy}, \bibinfo{person}{Wei Liu}, \bibinfo{person}{Yangqing Jia}, \bibinfo{person}{Pierre Sermanet}, \bibinfo{person}{Scott Reed}, \bibinfo{person}{Dragomir Anguelov}, \bibinfo{person}{Dumitru Erhan}, \bibinfo{person}{Vincent Vanhoucke}, {and} \bibinfo{person}{Andrew Rabinovich}.} \bibinfo{year}{2015}\natexlab{}.
\newblock \showarticletitle{Going deeper with convolutions}. In \bibinfo{booktitle}{\emph{Proceedings of the IEEE conference on computer vision and pattern recognition}}. \bibinfo{pages}{1--9}.
\newblock


\bibitem[Szegedy et~al\mbox{.}(2016)]%
        {szegedy2016rethinking}
\bibfield{author}{\bibinfo{person}{Christian Szegedy}, \bibinfo{person}{Vincent Vanhoucke}, \bibinfo{person}{Sergey Ioffe}, \bibinfo{person}{Jon Shlens}, {and} \bibinfo{person}{Zbigniew Wojna}.} \bibinfo{year}{2016}\natexlab{}.
\newblock \showarticletitle{Rethinking the inception architecture for computer vision}. In \bibinfo{booktitle}{\emph{Proceedings of the IEEE conference on computer vision and pattern recognition}}. \bibinfo{pages}{2818--2826}.
\newblock


\bibitem[Szegedy et~al\mbox{.}(2013)]%
        {szegedy2013intriguing}
\bibfield{author}{\bibinfo{person}{Christian Szegedy}, \bibinfo{person}{Wojciech Zaremba}, \bibinfo{person}{Ilya Sutskever}, \bibinfo{person}{Joan Bruna}, \bibinfo{person}{Dumitru Erhan}, \bibinfo{person}{Ian Goodfellow}, {and} \bibinfo{person}{Rob Fergus}.} \bibinfo{year}{2013}\natexlab{}.
\newblock \showarticletitle{Intriguing properties of neural networks}.
\newblock \bibinfo{journal}{\emph{arXiv preprint arXiv:1312.6199}} (\bibinfo{year}{2013}).
\newblock


\bibitem[Tan and Le(2021)]%
        {tan2021efficientnetv2}
\bibfield{author}{\bibinfo{person}{Mingxing Tan} {and} \bibinfo{person}{Quoc Le}.} \bibinfo{year}{2021}\natexlab{}.
\newblock \showarticletitle{Efficientnetv2: Smaller models and faster training}. In \bibinfo{booktitle}{\emph{International conference on machine learning}}. PMLR, \bibinfo{pages}{10096--10106}.
\newblock


\bibitem[Tang et~al\mbox{.}(2023)]%
        {tang2023adversarial}
\bibfield{author}{\bibinfo{person}{Guijian Tang}, \bibinfo{person}{Tingsong Jiang}, \bibinfo{person}{Weien Zhou}, \bibinfo{person}{Chao Li}, \bibinfo{person}{Wen Yao}, {and} \bibinfo{person}{Yong Zhao}.} \bibinfo{year}{2023}\natexlab{}.
\newblock \showarticletitle{Adversarial patch attacks against aerial imagery object detectors}.
\newblock \bibinfo{journal}{\emph{Neurocomputing}}  \bibinfo{volume}{537} (\bibinfo{year}{2023}), \bibinfo{pages}{128--140}.
\newblock


\bibitem[Team et~al\mbox{.}(2023)]%
        {team2023gemini}
\bibfield{author}{\bibinfo{person}{Gemini Team}, \bibinfo{person}{Rohan Anil}, \bibinfo{person}{Sebastian Borgeaud}, \bibinfo{person}{Yonghui Wu}, \bibinfo{person}{Jean-Baptiste Alayrac}, \bibinfo{person}{Jiahui Yu}, \bibinfo{person}{Radu Soricut}, \bibinfo{person}{Johan Schalkwyk}, \bibinfo{person}{Andrew~M Dai}, \bibinfo{person}{Anja Hauth}, {et~al\mbox{.}}} \bibinfo{year}{2023}\natexlab{}.
\newblock \showarticletitle{Gemini: a family of highly capable multimodal models}.
\newblock \bibinfo{journal}{\emph{arXiv preprint arXiv:2312.11805}} (\bibinfo{year}{2023}).
\newblock


\bibitem[Tu et~al\mbox{.}(2019)]%
        {tu2019autozoom}
\bibfield{author}{\bibinfo{person}{Chun-Chen Tu}, \bibinfo{person}{Paishun Ting}, \bibinfo{person}{Pin-Yu Chen}, \bibinfo{person}{Sijia Liu}, \bibinfo{person}{Huan Zhang}, \bibinfo{person}{Jinfeng Yi}, \bibinfo{person}{Cho-Jui Hsieh}, {and} \bibinfo{person}{Shin-Ming Cheng}.} \bibinfo{year}{2019}\natexlab{}.
\newblock \showarticletitle{Autozoom: Autoencoder-based zeroth order optimization method for attacking black-box neural networks}. In \bibinfo{booktitle}{\emph{Proceedings of the AAAI Conference on Artificial Intelligence}}, Vol.~\bibinfo{volume}{33}. \bibinfo{pages}{742--749}.
\newblock


\bibitem[Vaswani et~al\mbox{.}(2017)]%
        {vaswani2017attention}
\bibfield{author}{\bibinfo{person}{Ashish Vaswani}, \bibinfo{person}{Noam Shazeer}, \bibinfo{person}{Niki Parmar}, \bibinfo{person}{Jakob Uszkoreit}, \bibinfo{person}{Llion Jones}, \bibinfo{person}{Aidan~N Gomez}, \bibinfo{person}{{\L}ukasz Kaiser}, {and} \bibinfo{person}{Illia Polosukhin}.} \bibinfo{year}{2017}\natexlab{}.
\newblock \showarticletitle{Attention is all you need}.
\newblock \bibinfo{journal}{\emph{Advances in neural information processing systems}}  \bibinfo{volume}{30} (\bibinfo{year}{2017}).
\newblock


\bibitem[Wei et~al\mbox{.}(2022)]%
        {wei2022simultaneously}
\bibfield{author}{\bibinfo{person}{Xingxing Wei}, \bibinfo{person}{Ying Guo}, \bibinfo{person}{Jie Yu}, {and} \bibinfo{person}{Bo Zhang}.} \bibinfo{year}{2022}\natexlab{}.
\newblock \showarticletitle{Simultaneously optimizing perturbations and positions for black-box adversarial patch attacks}.
\newblock \bibinfo{journal}{\emph{IEEE transactions on pattern analysis and machine intelligence}} (\bibinfo{year}{2022}).
\newblock


\bibitem[Williams and Li(2023)]%
        {williams2023black}
\bibfield{author}{\bibinfo{person}{Phoenix~Neale Williams} {and} \bibinfo{person}{Ke Li}.} \bibinfo{year}{2023}\natexlab{}.
\newblock \showarticletitle{Black-Box Sparse Adversarial Attack via Multi-Objective Optimisation}. In \bibinfo{booktitle}{\emph{Proceedings of the IEEE/CVF Conference on Computer Vision and Pattern Recognition}}. \bibinfo{pages}{12291--12301}.
\newblock


\bibitem[Wu et~al\mbox{.}(2021)]%
        {wu2021genetic}
\bibfield{author}{\bibinfo{person}{Chenwang Wu}, \bibinfo{person}{Wenjian Luo}, \bibinfo{person}{Nan Zhou}, \bibinfo{person}{Peilan Xu}, {and} \bibinfo{person}{Tao Zhu}.} \bibinfo{year}{2021}\natexlab{}.
\newblock \showarticletitle{Genetic algorithm with multiple fitness functions for generating adversarial examples}. In \bibinfo{booktitle}{\emph{2021 IEEE Congress on Evolutionary Computation (CEC)}}. IEEE, \bibinfo{pages}{1792--1799}.
\newblock


\bibitem[Xu et~al\mbox{.}(2022)]%
        {xu2022groupvit}
\bibfield{author}{\bibinfo{person}{Jiarui Xu}, \bibinfo{person}{Shalini De~Mello}, \bibinfo{person}{Sifei Liu}, \bibinfo{person}{Wonmin Byeon}, \bibinfo{person}{Thomas Breuel}, \bibinfo{person}{Jan Kautz}, {and} \bibinfo{person}{Xiaolong Wang}.} \bibinfo{year}{2022}\natexlab{}.
\newblock \showarticletitle{Groupvit: Semantic segmentation emerges from text supervision}. In \bibinfo{booktitle}{\emph{Proceedings of the IEEE/CVF Conference on Computer Vision and Pattern Recognition}}. \bibinfo{pages}{18134--18144}.
\newblock


\bibitem[Yang et~al\mbox{.}(2021)]%
        {yang2021defending}
\bibfield{author}{\bibinfo{person}{Karren Yang}, \bibinfo{person}{Wan-Yi Lin}, \bibinfo{person}{Manash Barman}, \bibinfo{person}{Filipe Condessa}, {and} \bibinfo{person}{Zico Kolter}.} \bibinfo{year}{2021}\natexlab{}.
\newblock \showarticletitle{Defending multimodal fusion models against single-source adversaries}. In \bibinfo{booktitle}{\emph{Proceedings of the IEEE/CVF Conference on Computer Vision and Pattern Recognition}}. \bibinfo{pages}{3340--3349}.
\newblock


\bibitem[Yu and Koltun(2015)]%
        {yu2015multi}
\bibfield{author}{\bibinfo{person}{Fisher Yu} {and} \bibinfo{person}{Vladlen Koltun}.} \bibinfo{year}{2015}\natexlab{}.
\newblock \showarticletitle{Multi-scale context aggregation by dilated convolutions}.
\newblock \bibinfo{journal}{\emph{arXiv preprint arXiv:1511.07122}} (\bibinfo{year}{2015}).
\newblock


\bibitem[Zhong et~al\mbox{.}(2022)]%
        {zhong2022regionclip}
\bibfield{author}{\bibinfo{person}{Yiwu Zhong}, \bibinfo{person}{Jianwei Yang}, \bibinfo{person}{Pengchuan Zhang}, \bibinfo{person}{Chunyuan Li}, \bibinfo{person}{Noel Codella}, \bibinfo{person}{Liunian~Harold Li}, \bibinfo{person}{Luowei Zhou}, \bibinfo{person}{Xiyang Dai}, \bibinfo{person}{Lu Yuan}, \bibinfo{person}{Yin Li}, {et~al\mbox{.}}} \bibinfo{year}{2022}\natexlab{}.
\newblock \showarticletitle{Regionclip: Region-based language-image pretraining}. In \bibinfo{booktitle}{\emph{Proceedings of the IEEE/CVF Conference on Computer Vision and Pattern Recognition}}. \bibinfo{pages}{16793--16803}.
\newblock


\bibitem[Zhou et~al\mbox{.}(2021)]%
        {zhou2021data}
\bibfield{author}{\bibinfo{person}{Xingyu Zhou}, \bibinfo{person}{Zhisong Pan}, \bibinfo{person}{Yexin Duan}, \bibinfo{person}{Jin Zhang}, {and} \bibinfo{person}{Shuaihui Wang}.} \bibinfo{year}{2021}\natexlab{}.
\newblock \showarticletitle{A data independent approach to generate adversarial patches}.
\newblock \bibinfo{journal}{\emph{Machine Vision and Applications}}  \bibinfo{volume}{32} (\bibinfo{year}{2021}), \bibinfo{pages}{1--9}.
\newblock


\bibitem[Zhou et~al\mbox{.}(2023)]%
        {zhou2023advclip}
\bibfield{author}{\bibinfo{person}{Ziqi Zhou}, \bibinfo{person}{Shengshan Hu}, \bibinfo{person}{Minghui Li}, \bibinfo{person}{Hangtao Zhang}, \bibinfo{person}{Yechao Zhang}, {and} \bibinfo{person}{Hai Jin}.} \bibinfo{year}{2023}\natexlab{}.
\newblock \showarticletitle{Advclip: Downstream-agnostic adversarial examples in multimodal contrastive learning}. In \bibinfo{booktitle}{\emph{Proceedings of the 31st ACM International Conference on Multimedia}}. \bibinfo{pages}{6311--6320}.
\newblock


\end{thebibliography}
